\documentclass[10pt,twocolumn,letterpaper]{article}

\usepackage{iccv}
\usepackage{times}
\usepackage{epsfig}
\usepackage{graphicx}
\usepackage{amsmath}
\usepackage{amssymb}
\usepackage{caption}

\usepackage{blindtext}
\usepackage{booktabs}  %
\usepackage{multirow}  %
\usepackage{makecell}
\usepackage{float}
\usepackage{makecell}
\usepackage{pifont}
\usepackage{stfloats}
\DeclareMathOperator*{\argmin}{\arg\min}

\usepackage[symbol]{footmisc}

\usepackage{pdfpages} 
\usepackage{pgffor} 

\makeatletter
\AtBeginDocument{\let\LS@rot\@undefined}
\makeatother

\def\supplementfilename{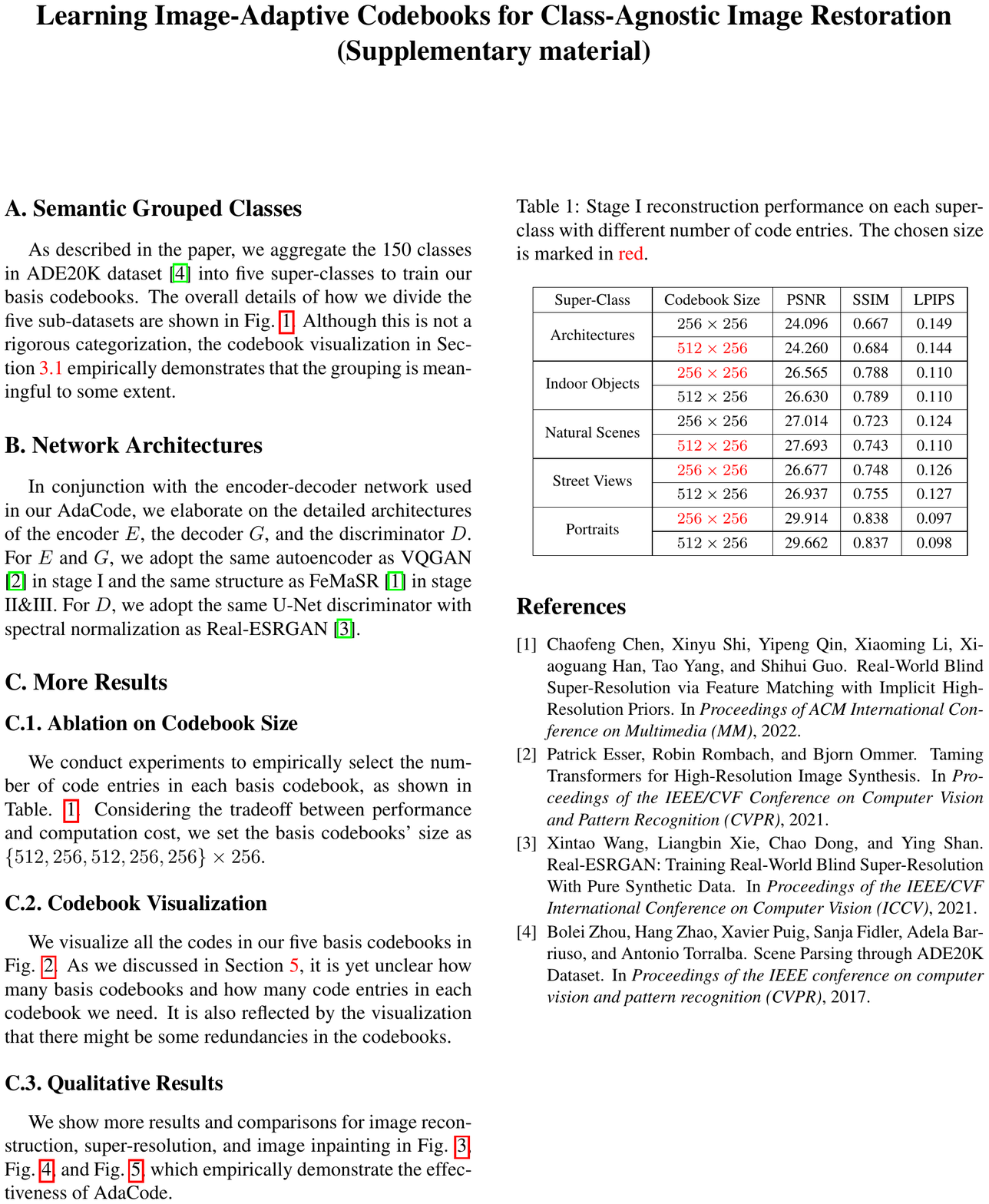}

\pdfximage{\supplementfilename}
\def\numbersupplementpages{\the\pdflastximagepages}

\newif\ifarXiv
\arXivtrue

\usepackage{hyperref}
\hypersetup{breaklinks=true,bookmarks=false}

\iccvfinalcopy %

\def\iccvPaperID{10266} %

\ificcvfinal\fi

\begin{document}

\title{Learning Image-Adaptive Codebooks for Class-Agnostic Image Restoration}

\author{Kechun Liu$^{1,3}$\footnote[1] 
\;\;\;\;\; Yitong Jiang$^{2}$ \;\; Inchang Choi$^{3}$ \;\; Jinwei Gu$^{2}$ \;\; \\ 
		$^1$ University of Washington  \;\;  
		$^2$ The Chinese University of Hong Kong  \;\; 
		$^3$ SenseBrain \;\;}

\ificcvfinal\thispagestyle{empty}\fi

\twocolumn[{%
\renewcommand\twocolumn[1][]{#1}%
\maketitle
\begin{center}
    \includegraphics[width=\textwidth]{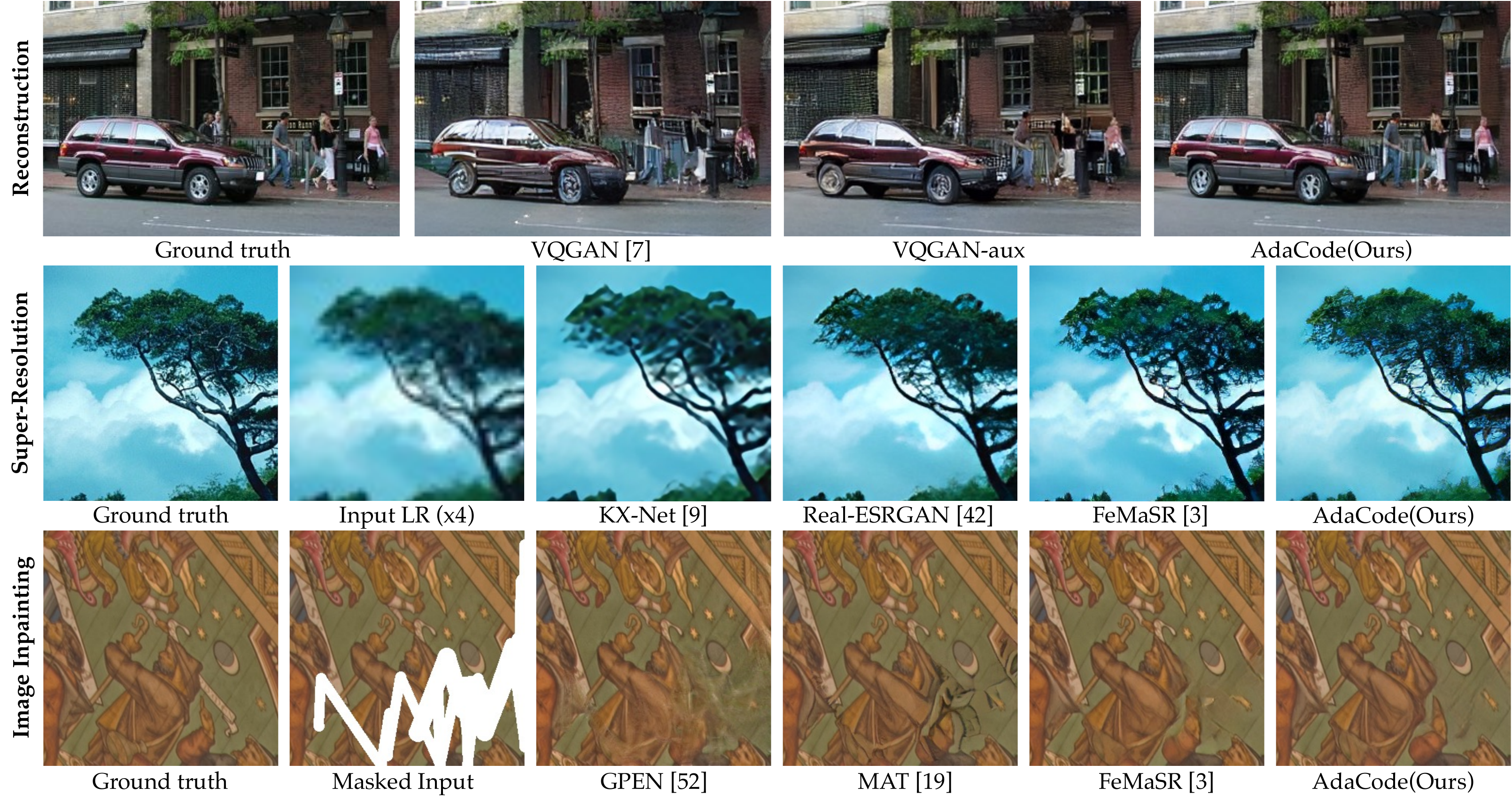}
    \captionof{figure}{We propose an image-adaptive codebook learning method, named \textbf{AdaCode}, for class-agnostic image restoration. For a number of image reconstruction and restoration tasks (e.g., super-resolution and inpainting), the proposed AdaCode achieved significantly better performance than latest prior work, such as VQGAN\cite{esser2021taming} and VQGAN with our merged basis codebooks (referred to as VQGAN-aux) for reconstruction; KX-Net\cite{fu2022kxnet}, Real-ESRGAN\cite{wang2021real} and FeMaSR\cite{chen2022real} for super-resolution; GPEN\cite{yang2021gan}, MAT\cite{li2022mat} and our trained FeMaSR\cite{chen2022real} for image inpainting.  (\textbf{Zoom in for best view})}
    \label{fig:representative}
\end{center}
}]
\footnotetext[1]{kechun@cs.washington.edu\\Work done during an internship at SenseBrain.}
\begin{abstract}
Recent work on discrete generative priors, in the form of codebooks, has shown exciting performance for image reconstruction and restoration, as the discrete prior space spanned by the codebooks increases the robustness against diverse image degradations. Nevertheless, these methods require separate training of codebooks for different image categories, which limits their use to specific image categories only (\eg face, architecture, \etc), and fail to handle arbitrary natural images. In this paper, we propose AdaCode for learning image-adaptive codebooks for class-agnostic image restoration. Instead of learning a single codebook for each image category, we learn a set of basis codebooks. For a given input image, AdaCode learns a weight map with which we compute a weighted combination of these basis codebooks for adaptive image restoration. Intuitively, AdaCode is a more flexible and expressive discrete generative prior than previous work. Experimental results demonstrate that AdaCode achieves state-of-the-art performance on image reconstruction and restoration tasks, including image super-resolution and inpainting. We will release the code and models upon publication.
\end{abstract}

\section{Introduction}
In recent years, discrete generative priors (in the form of codebooks) \cite{van2017neural, esser2021taming} have shown impressive performance for image synthesis \cite{esser2021taming, gu2022vector, wang2022restoreformer, zhou2022towards}, exhibiting reduced mode collapse and more stable training.
These learned codebooks essentially provide strong priors for compressing and reconstructing natural images, even in the presence of severe degradation. Nevertheless, these methods have a common limitation. The codebooks need to be learned separately for each image category (\eg, face, architecture), which restricts their applicability to arbitrary natural images \cite{esser2021taming, wang2022restoreformer, zhou2022towards}. Although FeMaSR \cite{chen2022real} attempted to learn a single general codebook for all image categories, the expressiveness of the codebook is limited by the complexity of natural images. For example, as shown in Fig. \ref{fig:representative}, an image often includes textural and structural contents from multiple categories (e.g., face, man-made structural edges, repetitive texture, natural texture). It is challenging to rely on a single universal codebook to capture all. Prior work such as VQGAN \cite{esser2021taming} and FeMaSR \cite{chen2022real} often introduce noticeable artifacts for image reconstruction and restoration.
\begin{figure}[h!]
    \centering
    \includegraphics[width=\columnwidth]{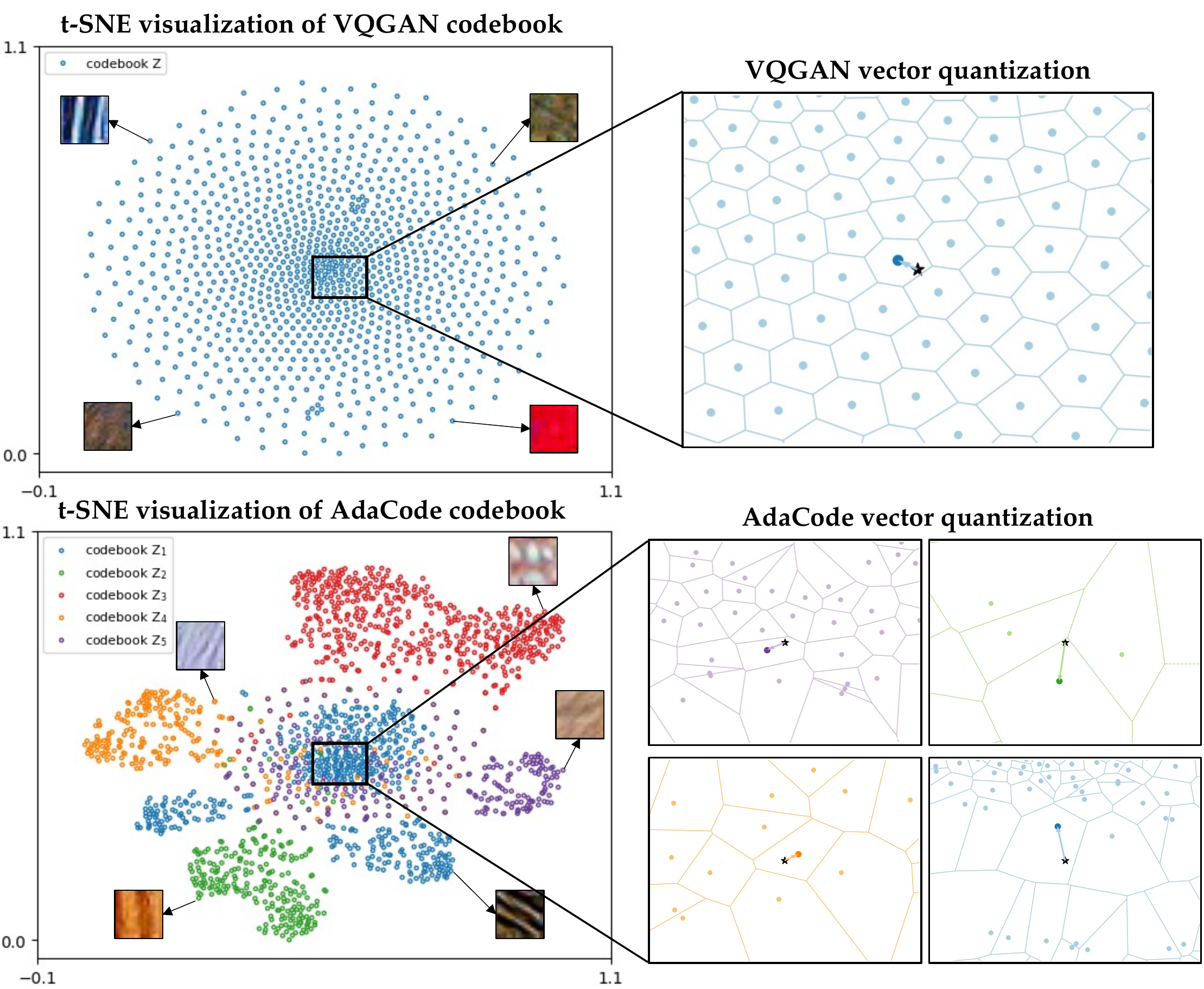}
    \caption{\textbf{Intuition of AdaCode vs a single codebook.} Top: Prior work VQGAN\cite{esser2021taming} uses a single codebook as the learned representation in the discrete latent space, which may not fully capture complex visual patterns. Bottom: AdaCode learns multiple basis codebooks, each representing a different discretization of the latent space corresponding to different visual appearances. For an input image, its latent representation is a weighted linear combination of the codes from these codebooks. AdaCode thus is a more flexible representation for class-agnostic image restoration.}
    \label{fig:toy}
\end{figure}

Is it possible to learn a class-agnostic discrete generative prior for image reconstruction and restoration? Inspired by a recent work \cite{zeng2020learning}, we propose AdaCode, which learns image-adaptive codebooks for class-agnostic image reconstruction and restoration. Instead of learning a single codebook for all categories of images, we learn a set of basis codebooks. For a given input image, AdaCode learns a weight map that determines the contribution of each basis codebook to the final representation. Intuitively, this design allows AdaCode to learn a more flexible and expressive discrete generative prior than previous work, as demonstrated in Fig.~\ref{fig:toy}. In contrast to VQGAN \cite{esser2021taming} and FeMaSR \cite{chen2022real}, which utilize a single partition for the latent space and assign each image feature an exclusive discrete representation, AdaCode learns various partitions to the latent space from different perspectives -- each corresponding to the learning of one of the basis codebooks. The discrete generative prior for an arbitrary image is a weighted linear combination derived from these basis codebooks, resulting in a more flexible and expressive representation. As depicted in Fig. \ref{fig:representative}, AdaCode outperforms previous work in various image restoration tasks, effectively preserving scene structure and texture.  

We evaluated AdaCode on both image reconstruction and image restoration tasks (i.e., super-resolution and image inpainting). Across multiple benchmark datasets, AdaCode achieved state-of-the-art performance, while maintaining a comparable codebook size and computational cost. We will release the code and model upon publication.

\section{Related Work}
\begin{figure*}[h!]
    \centering
    \includegraphics[width=\textwidth]{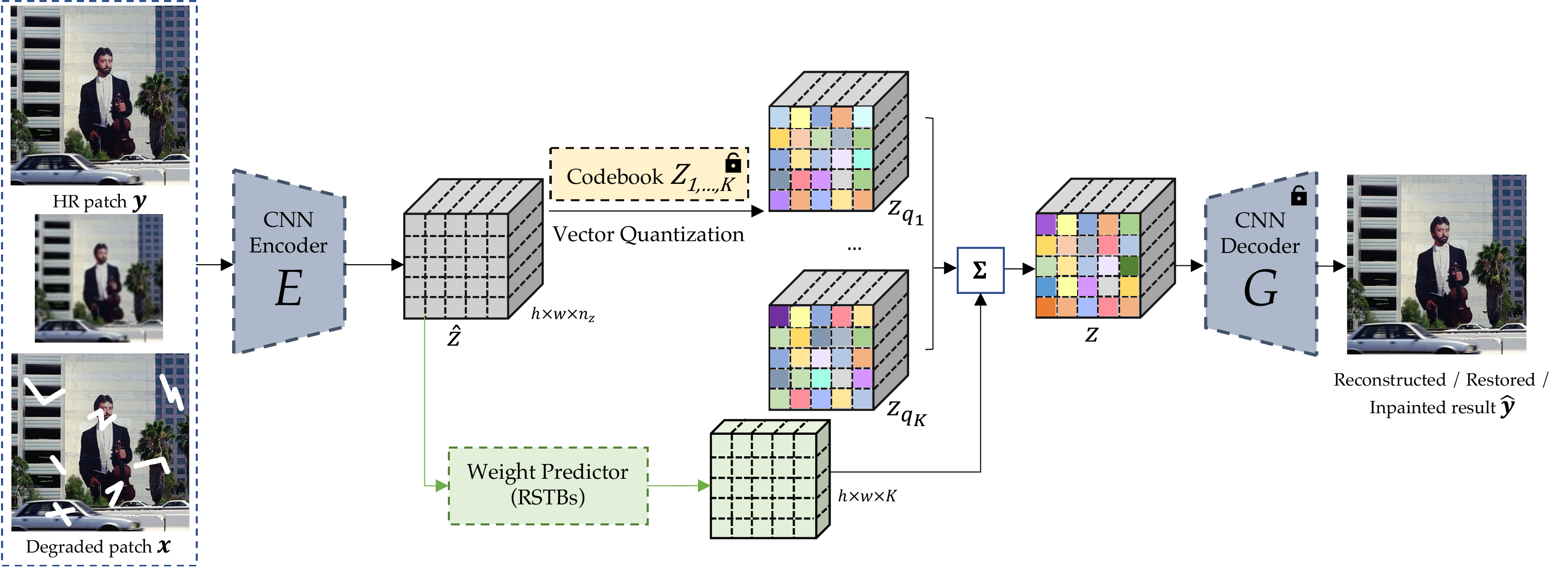}
    \caption{\textbf{The framework of the proposed AdaCode method.} The training of AdaCode incorporates three stages: Class-specific Codebook Pretraining, AdaCode Representation Learning, and Restoration via AdaCode. 
    The lock icon indicates that the codebooks $Z_{1,...,K}$ are fixed in Stage \uppercase\expandafter{\romannumeral2}\&\uppercase\expandafter{\romannumeral3}, and the decoder $G$ is fixed in Stage \uppercase\expandafter{\romannumeral3}.
    See Section. \ref{sec:method} for more details.
    }
    \label{fig:framework}
\end{figure*}

\paragraph{Visual Representation Dictionary Learning}
Learning representation dictionaries in visual understanding has demonstrated its great power in image restoration tasks such as super-resolution \cite{yang2010image, zhu2014fast}, denoising \cite{zhao2014hyperspectral}, and image inpainting \cite{fadili2009inpainting, shen2009image}. Using DNNs, VQVAE \cite{van2017neural} first introduces a generative autoencoder model that learns discrete latent representations, also known as ``codebook". The following VQGAN \cite{esser2021taming} employs perceptual and adversarial loss to train the visual codebook, resulting in better image generation quality with a relatively small codebook size. The representation dictionary-based generative model has inspired various impressive image generation work \cite{gu2022vector,zhou2022towards,yu2021vector, chen2022real, wang2022restoreformer}, as well as our AdaCode.

The use of dictionaries is not limited to image restoration. Referred as lookup tables (LUTs), the dictionaries are also applied to optimize color transforms \cite{zeng2020learning, liu20224d, yang2022adaint, yang2022seplut}. In 3D-LUT \cite{zeng2020learning}, multiple LUTs are learned to serve as the bases of the LUT space. And a CNN is trained to predict weights to fuse the bases into an image-adaptive LUT. Inspired by 3D LUT, we leverage the discrete codebooks from VQGAN as the bases of the image latent space to build our image-adaptive codebook, AdaCode. Such design allows our method to fully and flexibly exploit the latent codes to represent diverse and complex natural images.

\paragraph{DNN-based Image Restoration}
DNN has been widely used for image restoration tasks, \ie single image super-resolution (SISR) and image inpainting.
In the field of super-resolution (SR), recent studies focus on recovering low-resolution input images with unknown degradation types, which is more relevant to real-world scenarios. They restore the images either by learning the degradation representations \cite{ren2020neural,liang2021flow,wang2021unsupervised,wang2021real,zhang2021designing,fu2022kxnet}, or by training unified generative networks with LR-HR pairs \cite{ji2020real, wei2020component, rad2021benefiting}. Some studies introduce additional image priors, such as latent representations \cite{menon2020pulse, chan2021glean, pan2021exploiting} and discrete codebooks \cite{esser2021taming, zhou2022towards, chen2022real}, to address unrealistic textures or over-smoothed areas commonly observed in GAN-based methods. However, a single partition of the latent space is often insufficient to model the intricate patterns in natural images, resulting in specific textures being generated regardless of the image content.

In the case of image inpainting, researchers often leverage deep generative models to fill missing image regions with plausible content \cite{pathak2016context,liu2020rethinking,wang2018image,zeng2019learning,deng2018uv}. Some approaches incorporate additional discriminators \cite{satoshi2017globally}, partial or gated convolutions \cite{liu2018image, yu2019free}, semantic texture or context \cite{nazeri2019edgeconnect,xiong2019foreground, ren2019structureflow, guo2021image, song2018spg, hong2018learning}, or transformers \cite{li2022mat, dong2022incremental,liu2022reduce,wan2021high} to enhance the quality of inpainted results. However, these methods often require separate experiments for different image patterns, \ie natural scenes, faces, \etc, due to the significant variations among them.

Both SR and inpainting methods face the challenge of effectively modeling complex visual patterns with a single model or codebook. In our proposed approach, AdaCode, we address this challenge by leveraging adaptive codebooks, enabling realistic and robust restoration results for general images.

\section{Methodology}
\label{sec:method}
Building upon VQGAN \cite{esser2021taming}, we introduce an additional degree of freedom to the codebook and construct an adaptive codebook (AdaCode) to model the intricate high-resolution natural image patterns. We summarize the AdaCode model in Fig.\ref{fig:framework}. Our method's overall training consists of three sequential stages. In the first stage (Sec \ref{sec:stage1}), we divide our HQ dataset into multiple semantic subsets and train a class-specific VQGAN \cite{esser2021taming} on each subset. In the second stage, using the fixed pretrained class-specific codebooks as bases, we leverage a transformer block to generate weight maps and train the AdaCode through the self-reconstruction task (Sec \ref{sec:stage2}). In the last stage (Sec \ref{sec:stage3}), we employ the AdaCode with fixed codebooks and fixed image decoder to address downstream restoration tasks, \eg Super-Resolution and Image Inpainting. The details of each stage are discussed in the following sections.

\subsection{Codebook Pretraining (Stage \uppercase\expandafter{\romannumeral1})}
\label{sec:stage1}

\begin{figure}[h!]
\centering
\includegraphics[width=0.95\columnwidth]{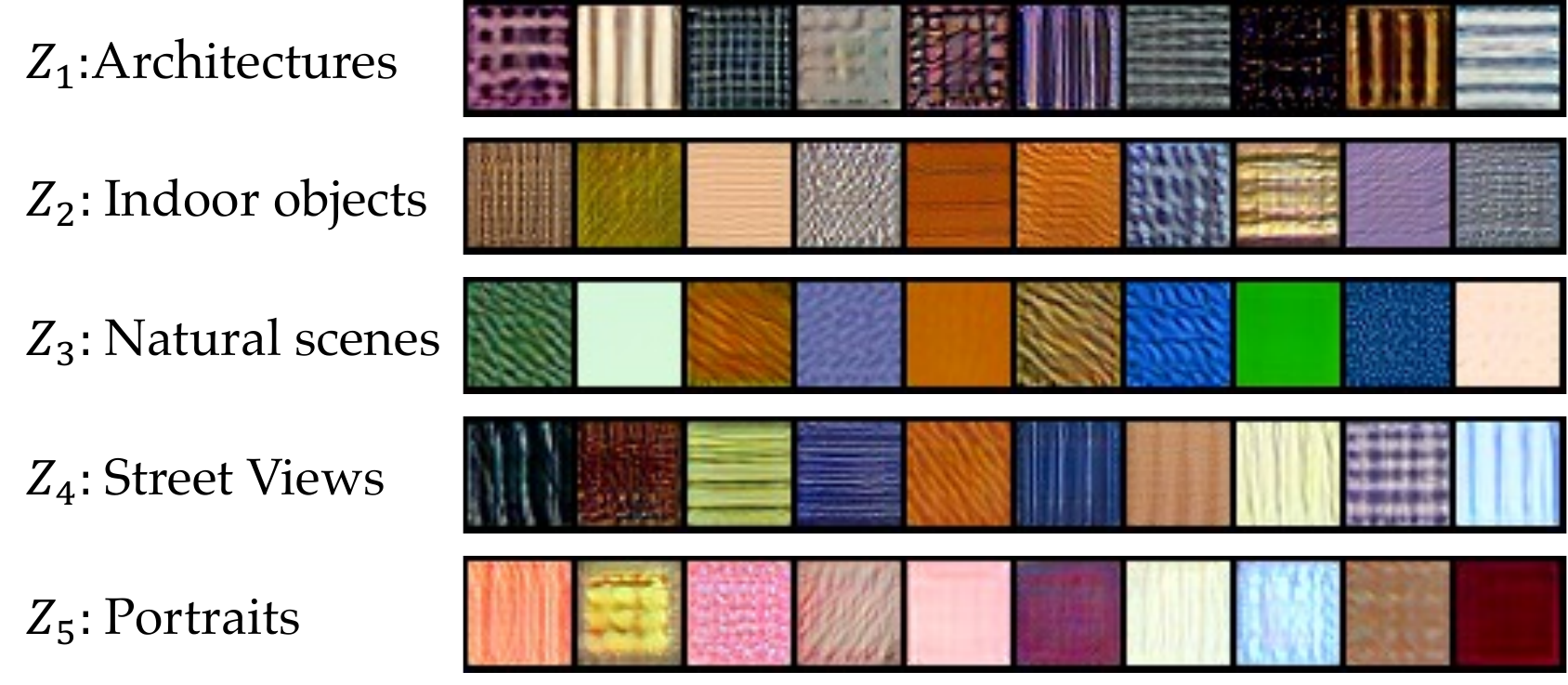}
  \caption{\textbf{Visualization of the learned codes for different categories of visual appearance.} We randomly sample 10 code entries from each codebook and visualize them by projecting a tiled $4\times4$ feature map onto a $32\times32$ texture patch using the corresponding decoder. As expected, the codes from different categories exhibit distinct features. }
  \label{fig:codebook}
\end{figure}

\paragraph{Diversify Basis Codebooks}
To enhance the expressiveness of AdaCode, we aim to diversify the basis codebooks. Rather than applying various initializations as 3D-LUT \cite{zeng2020learning}, we achieve codebook divergence by training them on different HR subsets of our dataset. To accomplish this, we utilize an off-the-shelf SegFormer model \cite{xie2021segformer} to perform semantic segmentation with 150 classes from the ADE20K dataset \cite{zhou2017scene}. Each HR patch is labeled according to the semantic class with the largest area. We then group the 150 classes into 5 super-classes: Architectures, Indoor objects, Natural scenes, Street views, Portraits, and obtain the 5 semantic HR subsets accordingly. It is worth noting that the separation of subsets is not rigorous and each subset may contain semantic contexts from other subsets, as most images contain multiple semantics. More details regarding the clustering of the 150 classes can be found in the Appendix. This approach not only diversifies the codebooks but also provides various ways to partition the latent feature space. We randomly visualize some code samples in Fig. \ref{fig:codebook}.

\paragraph{Learning Basis Codebooks}
Given a HQ subset of class $k$, we train a quantized autoencoder to learn the class-specific basis codebook. As shown in Fig. \ref{fig:framework}, the input HR patch $y \in \mathbb{R}^{H\times W\times 3}$ is first passed through the encoder $E$ to generate the embedding $\hat{z} = E(y) \in \mathbb{R}^{h\times w\times n_z}$. Following VQVAE \cite{van2017neural} and VQGAN \cite{esser2021taming}, each entry $\hat{z}_i \in \mathbb{R}^{n_z}$ in $\hat{z}$ is replaced with its nearest code in the learnable codebook $Z_k \in \mathbb{R}^{N\times n_z}$ to construct the quantized embedding $z_{q_k}$:
\begin{equation}
\label{eqn:vq}
    z_{q_k} = \argmin_{c\in \{0,...,N-1\}}\vert\vert \hat{z} - Z_{k,c}\vert\vert
\end{equation}
where N is the number of codes in the corresponding codebook, $z_{q_k}$ denotes the quantized representation using $Z_k$, and $Z_{k,c}$ represent the $c$-th entry in codebook $Z_k$. After the feature quantization, the decoder $G$ reconstructs the HR patch $\hat{y}$ using $z_{q_k}$:
\begin{equation}
    \hat{y}=G(z_{q_k})\approx y
\end{equation}
The adversarial learning scheme is employed to train the encoder $E$, codebook $Z$, and decoder $G$ with the discriminator $D$. The detailed architectures of $E$, $D$, and $G$ are provided in the Appendix.

\paragraph{Training Objective}
To train the quantized autoencoder, we adopt 3 image-level losses: L1 loss $\mathcal{L}_1$, perceptual loss $\mathcal{L}_{per}$ \cite{johnson2016perceptual}, and adversarial loss $\mathcal{L}_{adv}$ \cite{goodfellow2014generative}, which are calculated using $\hat{y}$ and $y$.

Since the quantization in Eqn. \ref{eqn:vq} is non-differentiable, we adopt the straight-through gradient estimator in \cite{van2017neural, esser2021taming}, which directly copies the gradients from decoder $G$ to encoder $E$, enabling back-propagation and allowing end-to-end training using the code-level loss function $\mathcal{L}_{VQ}$:
\begin{equation}
    \mathcal{L}_{VQ}(E,G,Z_k) = \vert\vert sg[\hat{z}] - z_q\vert\vert^2_2 + \beta \cdot \vert\vert\hat{z} - sg[z_q]\vert\vert^2_2
\end{equation}
where $sg[\cdot]$ denotes the stop-gradient operation and $\beta=0.25$ is a hyper-parameter to control the update frequency of the codebook.

To further reinforce the semantics in the latent codebook and improve the texture restoration \cite{wang2018recovering}, we incorporate a VGG19-based regularization term $\mathcal{L}_{sem}$ into the codebook training process, following the approach in \cite{chen2022real}.:
\begin{equation}
    \mathcal{L}_{sem} = \vert\vert CONV(\hat{z}) - \Phi(y_k')\vert\vert^2_2
\end{equation}
where $\Phi$ denotes the feature extractor of VGG19 \cite{simonyan2014very}, and $CONV$ denotes a single convolutional layer to match the dimension of $\hat{z}$ and $\Phi(y_k')$.

With the above image-level and code-level losses, we can summarize the training objective in Stage \uppercase\expandafter{\romannumeral1} as:
\begin{equation}
    \mathcal{L}_{stage1}= \mathcal{L}_1 + \mathcal{L}_{per} + \lambda \cdot \mathcal{L}_{adv} + \mathcal{L}_{VQ} + \lambda \cdot \mathcal{L}_{sem}
\end{equation}
where the loss weight $\lambda$ is set to 0.1.

\subsection{AdaCode Representation Learning (Stage \uppercase\expandafter{\romannumeral2})}
\label{sec:stage2}

\begin{figure}[h!]
    \centering
    \includegraphics[width=0.95\columnwidth]{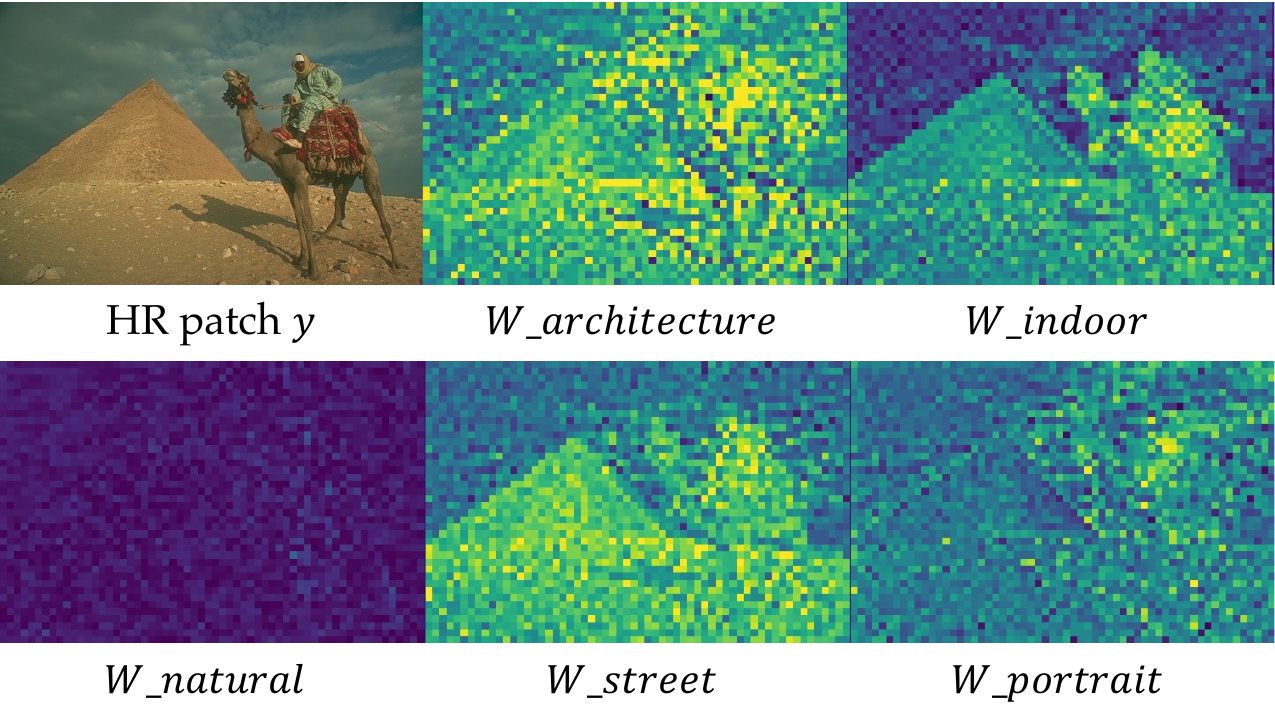}
    \caption{\textbf{An example showing the learned weight maps.} The input image contains multiple semantically-meaningful content (\eg pyramid, person, animal, sky) which cannot be well represented with a single codebook. Instead, AdaCode uses multiple basis codebooks and weight maps for discrete representations. As shown, the weight maps correlate to the semantics to some extent.}
    \label{fig:weight}
\end{figure}

Given the pretrained class-specific basis codebooks $Z_k, k\in\{1,...,K\}$, the latent feature space can be partitioned into non-overlapping cells in $K$ different ways. Specifically, for a given input HR patch $y$, $K$ quantized representations can be generated. Each distinct quantized representation $z_{q_k}$ obtains its code token from its corresponding semantic codebook. To combine the discrete representation $z_{q_{1,...,K}}$ into the AdaCode representation $z$, we employ a weight predictor module, which generates a $K$-channel weight map $w \in \mathbb{R}^{h\times w\times K}$, as illustrated in Fig. \ref{fig:framework} and Fig. \ref{fig:weight}. The weight predictor module consists of four residual swin transformer blocks (RSTBs) \cite{liang2021swinir} and a convolution layer to match the channels of weight map and $K$. $z$ is computed following Eqn. \ref{eqn:linearcombination}. Finally, the adaptive feature $z$ is reconstructed to HR patch $\hat{y}$ via the decoder $G$.
\begin{equation}
\label{eqn:linearcombination}
    z = \sum_{i} w_i \times z_{q_i}
\end{equation}

To efficiently train AdaCode and maintain a comparable number of parameters in the codebooks to VQGAN \cite{esser2021taming} and FeMaSR \cite{chen2022real}, which both set the codebook dimension to be $1024\times 512$, we set each of our class-specific codebooks to be $256\times 256$ or $512\times 256$. These codebooks are fixed during the training of stage \uppercase\expandafter{\romannumeral2} while the rest of the model, \ie the encoder $E$, the weight predictor, the decoder $G$, and the discriminator $D$, are trained using the objective in Eqn. \ref{eqn:Lstage2}. Each term  in this equation is defined in Sec. \ref{sec:stage1}.
\begin{equation}
\label{eqn:Lstage2}
    \mathcal{L}_{stage2} = \mathcal{L}_1 + \mathcal{L}_{per} + \lambda \cdot \mathcal{L}_{adv} + \mathcal{L}_{VQ}(E,G)
\end{equation}

\subsection{Restoration via AdaCode (Stage \uppercase\expandafter{\romannumeral3})}
\label{sec:stage3}
With the powerful decoder $G$, various image restoration tasks can be turned into a feature refinement problem through AdaCode scheme. From the perspective of latent space partition, each representation of the degraded input $x$ is pulled towards its nearest HR code entry, allowing for the information loss in $x$ to be relatively compensated. 
In comparison to the quantized representation $z_q$ using only one general codebook, the combination of $z_{q_{1,...,K}}$ with the weight map can be considered as adding an offset to $z_q$. The offset helps to alleviate the discontinuity among the discrete codes, which is demonstrated in our ablation study. To further showcase the effectiveness of AdaCode, we train our model on two ill-posed problems, \ie Single Image Super-Resolution and Image Inpainting.

In super-resolution and image inpainting tasks, the mapping between the input and the HR output has more than one solutions. Despite the benefits of the AdaCode scheme, restoring damaged content or missing details remains challenging given the uncertain degradations and the diversity of natural images. To better account for the degradation and improve the gradient flow, we adopt the encoder design in \cite{chen2022real} which utilizes a feature extraction module and a residual shortcut module during stage \uppercase\expandafter{\romannumeral3}.

Thanks to the excellent reconstruction model in stage \uppercase\expandafter{\romannumeral2}, given a Degraded-HR image pair, we can obtain the groundtruth representation $z_{gt}$ via the fixed model. Since the decoder $G$ is fixed in this stage, the restoration problem can be formulated as minimizing the distance between HR feature $z_{gt}$ and degraded feature $z$. To achieve this, we use a code-level loss that includes the InfoNCE loss in \cite{oord2018representation} and the style loss in \cite{johnson2016perceptual}. Following the design as SimCLR\cite{chen2020simple}, given a degraded image feature $z$, we use the HR feature $z_{gt}$ as the positive sample, while other $z_{gt}$ and $z$ from different source images in the same batch are treated as the negative samples. The code-level loss is defined as follows.
\begin{equation}
\begin{split}
    \mathcal{L}_{code} = \mathcal{L}_{\text{InfoNCE}}(z_{gt}, z) &+ \mathcal{L}_{style}(z_{gt}, z) \\&+ \beta \cdot \vert\vert\hat{z} - sg[z_{gt}]\vert\vert^2_2
\end{split}
\end{equation}
And the overall loss is summarized as:
\begin{equation}
    \mathcal{L}_{stage3} = \mathcal{L}_1 + \mathcal{L}_{per} + \lambda \cdot \mathcal{L}_{adv} + \mathcal{L}_{code}
\end{equation}

\section{Experiments}

\begin{figure*}[h!]
    \centering
    \includegraphics[width=0.92\textwidth]{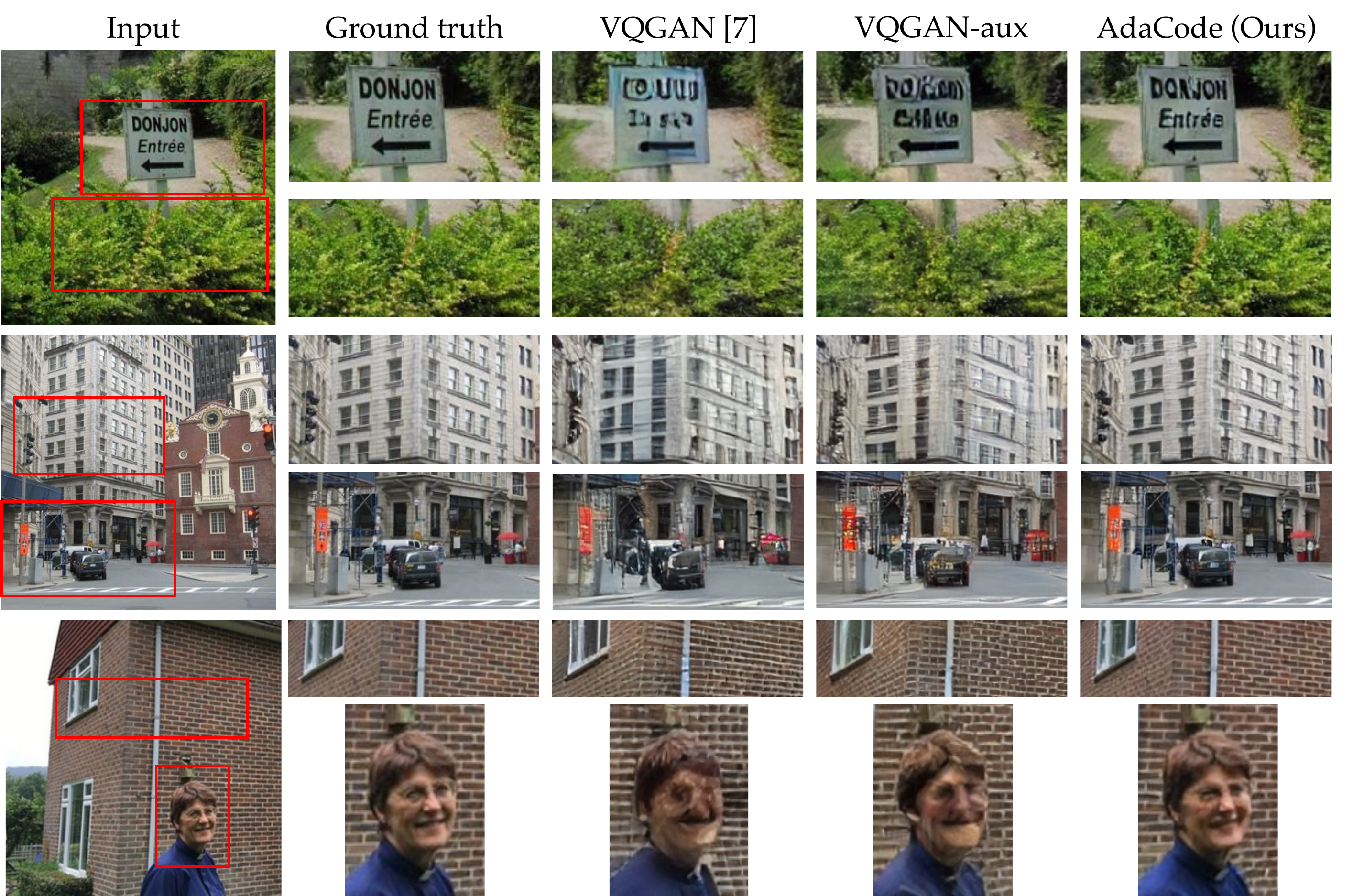}
    \caption{\textbf{Qualitative comparisons on image reconstruction task.} Various scenes and semantics, \ie text, plants, buildings, streets, and faces, are covered in these examples. Benefited from the adaptive codebook, our AdaCode is able to reconstruct images with higher quality and fidelity. (\textbf{Zoom in for best view})}
    \label{fig:recon}
    \vspace{-4pt}
\end{figure*}

\paragraph{Datasets}
\label{sec:dataset}
Our training dataset includes images from DIV2K train set \cite{Agustsson_2017_CVPR_Workshops}, Flickr2K \cite{lim2017enhanced}, DIV8K train set \cite{gu2019div8k}, and 10,000 face images from FFHQ \cite{karras2017progressive}. We generate the training patches by cropping images into non-overlapping patches at $512 \times 512$ resolution (face images in FFHQ are randomly resized with scale factors between [0.5, 1.0] before cropping). We adopt the same degradation model as BSRGAN \cite{zhang2021designing} to generate LR patches. The final training dataset consists of 198,061 patches. 

In the test stage, we evaluate the reconstruction task on OST dataset \cite{wang2018recovering}, which contains 300 images with rich textures. For super-resolution, we evaluate the performance on five classical benchmarks, \ie Set5, Set14, BSD100, Urban100, and Manga109, with $\times2$ and $\times4$ scales. For image inpainting, we apply a publicly available script \cite{yang2021gan} to randomly draw irregular polyline masks and generate masked images. The inpainting performance is evaluated on the validation sets of DIV2K \cite{Agustsson_2017_CVPR_Workshops} and DIV8K \cite{gu2019div8k}.

\paragraph{Evaluation Metrics}
For reconstruction, we adopt PSNR and SSIM as the evaluation metrics. For super-resolution, we employ an additional well-known perceptual score, LPIPS \cite{zhang2018unreasonable}. For image inpainting task, we use PSNR, LPIPS and a widely-used non-reference metric, FID \cite{heusel2017gans}.

\paragraph{Inplementation Details}
According to the size of each semantic dataset, we empirically set the codebook bases sizes to be $\{512, 256, 512, 256, 256\}\times 256$ for Architectures, Indoor Objects, Natural Scenes, Street Views, and Portraits.  For all stages, we represent the input image 
as a $32 \times 32$ code sequence.
We train each stage for 350k iterations with an Adam optimizer and a batch size of 32. The learning rates for the generator and discriminator are fixed as 1e-4 and 4e-4 separately. Our method is implemented with PyTorch and trained with 4 NVIDIA Tesla V100 GPUs.

\subsection{Expressiveness of AdaCode}
The key design in our work is to leverage the class-specific basis codebooks to construct an adaptive codebook, which supports more expressiveness even with a smaller codebook size. To verify our method's superiority, we evaluate the reconstruction performances using three codebook and model settings: (1) VQGAN \cite{esser2021taming} with its single general codebook. (2) VQGAN with a merged codebook concatenating all the basis codebooks, referred to as VQGAN-aux. (3) AdaCode with our adaptive codebook. The three approaches are trained with the same dataset (see Section. \ref{sec:dataset}) to guarantee a fair comparison.

\begin{table}[h!]
    \caption{\textbf{Quantitative comparison of reconstruction performance.} PSNR/SSIM$\uparrow$: the higher, the better.}\label{tab:recon}
    \centering
    \small
    \setlength\tabcolsep{2pt}
    \def\arraystretch{1.2}
    \begin{tabular}{|>{\centering\arraybackslash\hspace{0pt}}m{0.35\columnwidth}|>{\centering\arraybackslash\hspace{0pt}}m{0.3\columnwidth}|cc|}\hline
      \multirow{2}{*}{Method} & \multirow{2}{*}{\makecell[c]{Overall\\Codebook Size}} & \multicolumn{2}{c|}{Performance}\\ 
      \cline{3-4} & & PSNR & SSIM\\
      \cline{1-4} \textbf{VQGAN}\cite{chen2022real} & $1024 \times 512$ & 21.3557 & 0.5664\\
      \cline{1-4} \textbf{VQGAN-aux} & $1792 \times 256$ & 21.9219 & 0.6030\\
      \cline{1-4} \textbf{AdaCode (Ours)} & $1792 \times 256$ & \textbf{25.7629} & \textbf{0.7705}\\
      \hline
    \end{tabular}
\end{table}

As shown in Table. \ref{tab:recon}, our AdaCode obtains overwhelming reconstruction results with a comparable or smaller codebook size. The gap between (1) and (2) certifies that training class-specific codebooks helps the codes to capture more image textures, while the great improvement between (2) and (3) justifies the expressiveness facilitated by our adaptive codebook design. Fig. \ref{fig:recon} shows multiple scenarios, including plants, buildings, streets, portraits, and text which has distinct patterns but does not have a corresponding codebook in our experiments. AdaCode achieves exceedingly excellent results in all semantic cases, producing realistic and fidelitous reconstruction results.

\begin{figure*}
    \centering
    \includegraphics[width=\textwidth]{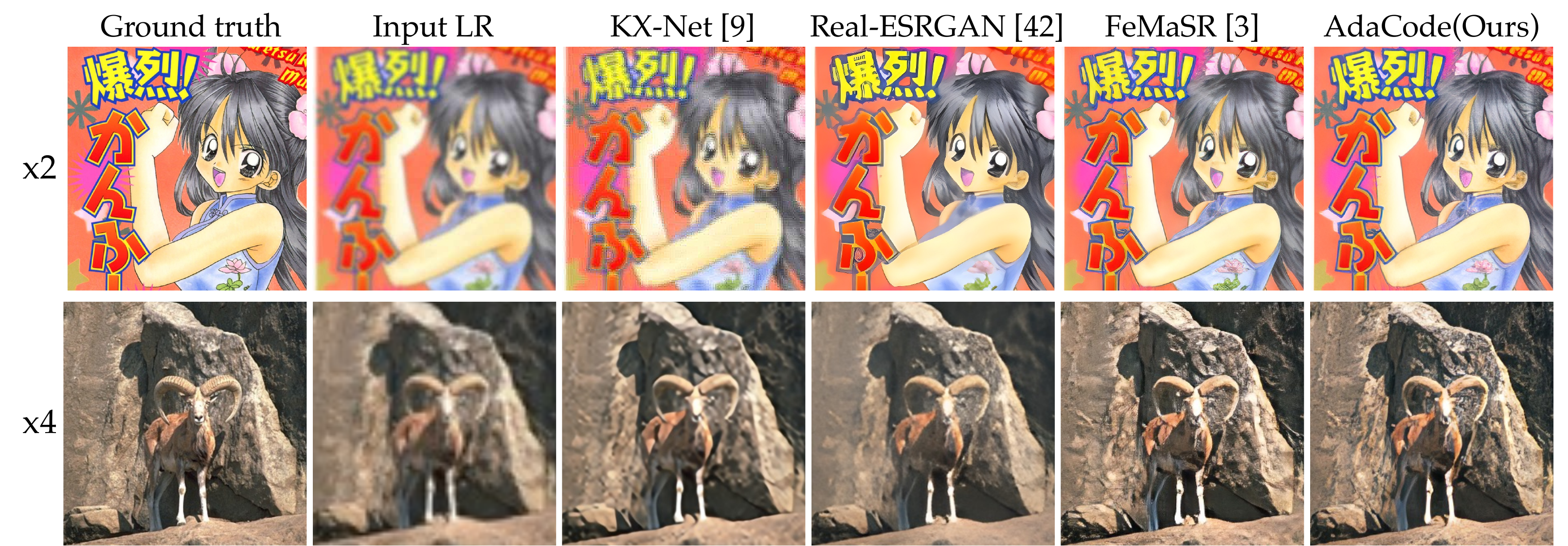}
    \caption{\textbf{Qualitative comparisons on super-resolution task with $\times2$ and $\times4$ upscale factor.} AdaCode restores LR with realistic and faithful details, while the competitive work either fails to deblur the LR, \ie KX-Net \cite{fu2022kxnet}, or generates artifacts or over-smooth areas,\ie Real-ESRGAN \cite{wang2021real} and FeMaSR \cite{chen2022real}. See Appendix for more results. (\textbf{Zoom in for best view})} 
    \label{fig:sr_result}
\end{figure*}

\setlength{\tabcolsep}{3.5pt}
\begin{table*}
\caption{\textbf{Quantitative comparison with state-of-the-art SISR methods.} PSNR/SSIM$\uparrow$: the higher, the better; LPIPS$\downarrow$: the lower, the better. The best and second best performance are marked in \textcolor{red}{red} and \textcolor{blue}{blue}.}
\label{tab:SISR}

\centering
\footnotesize
\def\arraystretch{1.3}
\begin{tabular}{|c|c|ccc|ccc|ccc|ccc|ccc|}\hline
  \multirow{2}{*}{Method} & \multirow{2}{*}{Scale} & \multicolumn{3}{c|}{Set5} & \multicolumn{3}{c|}{Set14} & \multicolumn{3}{c|}{BSDS100} & \multicolumn{3}{c|}{Urban100} & \multicolumn{3}{c|}{Manga109}\\ 
  \cline{3-17} & & PSNR & SSIM & LPIPS & PSNR & SSIM & LPIPS & PSNR & SSIM & LPIPS & PSNR & SSIM & LPIPS & PSNR & SSIM & LPIPS \\
  \cline{1-17} FeMaSR & x2 & 27.513 & 0.8250 & 0.1054 & 25.404 & 0.7565 & \textcolor{blue}{0.1292} & 24.907 & 0.7260 & \textcolor{blue}{0.1449} & 22.713 & 0.7573 & 0.1102 & \textcolor{blue}{21.584} & \textcolor{blue}{0.6938} & \textcolor{blue}{0.2270} \\
  \cline{1-17} KX-Net & x2 & 27.837 & \textcolor{blue}{0.8884} & \textcolor{blue}{0.0859} & 25.929 & \textcolor{blue}{0.8291} & 0.1301 & 25.818 & \textcolor{red}{0.8092} & 0.1734 & 22.675 & \textcolor{blue}{0.8016} & \textcolor{red}{0.0972} &  19.561 & 0.5049 & 0.5312\\
  \cline{1-17} Real-ESRGAN & x2 & \textcolor{red}{30.032} & \textcolor{red}{0.8930} & 0.1051 & \textcolor{red}{27.019} & \textcolor{red}{0.8379} & 0.1382 & \textcolor{red}{26.670} & \textcolor{blue}{0.7916} & 0.1526 & \textcolor{blue}{23.577} & \textcolor{red}{0.8115} & \textcolor{blue}{0.0976} & 20.815 & 0.6817 & 0.3056 \\
  \cline{1-17} \textbf{Ours} & x2 & \textcolor{blue}{29.213} & 0.8444 & \textcolor{red}{0.0844} & \textcolor{blue}{26.249} & 0.7702 & \textcolor{red}{0.1179} & \textcolor{blue}{26.038} & 0.7529 & \textcolor{red}{0.1374} & \textcolor{red}{23.663} & 0.7894 & 0.1009 & \textcolor{red}{22.023} & \textcolor{red}{0.7029} & \textcolor{red}{0.2150}\\
  \hline
  \cline{1-17} FeMaSR & x4 &  24.039 & 0.7452 & \textcolor{red}{0.1500} &  22.724 & 0.6354 & \textcolor{blue}{0.2045} &  21.957 & 0.5819 & \textcolor{blue}{0.2517} & 20.509 & 0.6426 & \textcolor{red}{0.1983} & 18.131 & \textcolor{blue}{0.5757} & \textcolor{blue}{0.3449} \\
  \cline{1-17} KX-Net & x4 & 21.922 & 0.7050 & 0.2080 & 21.182 & 0.6126 & 0.3044 & 21.708 & 0.5768 & 0.3972 & 18.603 & 0.5576 & 0.2704 & \textcolor{blue}{18.181} & 0.5323 & 0.6909\\
  \cline{1-17} Real-ESRGAN & x4 & \textcolor{blue}{25.263} & \textcolor{blue}{0.7665} & 0.1710 & \textcolor{blue}{24.100} & \textcolor{red}{0.7004} & 0.2338 & \textcolor{red}{23.776} & \textcolor{red}{0.6261} & 0.2819 & \textcolor{blue}{21.351} & \textcolor{red}{0.6625} & 0.2140 & \textcolor{red}{18.222} & \textcolor{red}{0.5917} & 0.4091\\
  \cline{1-17} \textbf{Ours} & x4 & \textcolor{red}{25.868} & \textcolor{red}{0.7731} & \textcolor{blue}{0.1505} & \textcolor{red}{24.158} & \textcolor{blue}{0.6662} & \textcolor{red}{0.2031} & \textcolor{blue}{23.129} & \textcolor{blue}{0.6041} & \textcolor{red}{0.2485} & \textcolor{red}{21.446} & \textcolor{blue}{0.6568} & \textcolor{blue}{0.2007} & 18.145 & 0.5664 & \textcolor{red}{0.3425}\\
  \cline{1-17}
\end{tabular}
\end{table*}

\subsection{Benchmarking Image Restoration Results}

\begin{figure*}
    \centering
    \includegraphics[width=\textwidth]{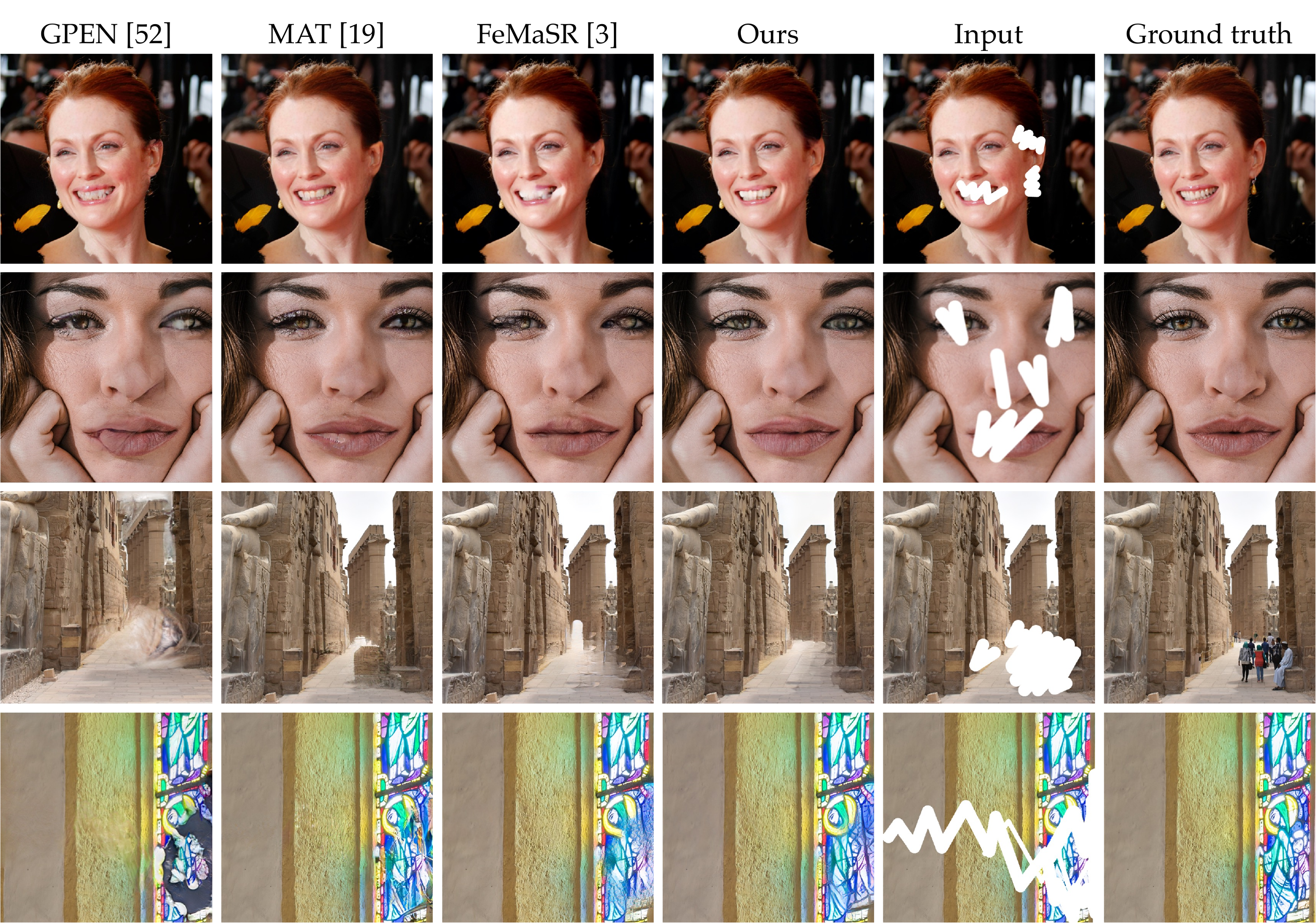}
    \caption{\textbf{Qualitative comparisons on image inpainting task.} The first two rows demonstrate the inpainting results of face images, while the last two rows show the recovered results of place images. See Appendix for more results.} 
    \label{fig:inpainting_result}
\end{figure*}

\paragraph{Super-Resolution}
We compare AdaCode with state-of-the-art models for Image Super-Resolution, including KX-Net\cite{fu2022kxnet}, Real-ESRGAN \cite{wang2021real}, and FeMaSR\cite{chen2022real}. Specifically, KX-Net iteratively learns the degradation kernels from the LR images; Real-ESRGAN learns super-resolution using pure synthetic data with high-order degradation model; FeMaSR utilizes a single perceptually rich codebook to restore the images. We use the original codes and weights from each method's official public repository to conduct comparisons, as shown in Table. \ref{tab:SISR} and Fig. \ref{fig:sr_result}.

\paragraph{Image Inpainting}
We compare AdaCode with state-of-the-art inpainting methods GPEN \cite{yang2021gan} and MAT \cite{li2022mat}.  To conduct a fair comparison, we retrain MAT on our training dataset as discussed in Section. \ref{sec:dataset}. Moreover, we train FeMaSR \cite{chen2022real} for this task to demonstrate the effectiveness of our adaptive codebooks over the single codebook in FeMaSR. As shown in Table. \ref{tab_inpainting}, AdaCode achieves state-of-the-art performance on various metrics. Qualitative comparisons in Fig. \ref{fig:inpainting_result} also illustrate that AdaCode consistently produce high-quality inpainting results across a wide range of scenes with a single model.

\setlength{\tabcolsep}{3.5pt}
\begin{table}
    \caption{\textbf{Quantitative comparison with state-of-the-art inpainting methods.} PSNR$\uparrow$: the higher, the better; LPIPS/FID$\downarrow$: the lower, the better. The best and second best performance are marked in \textcolor{red}{red} and \textcolor{blue}{blue}.}\label{tab_inpainting}
    \centering
    \footnotesize
    \def\arraystretch{1.3}
    \scalebox{1.1}{
    \begin{tabular}{|c|cc|cc|c|}\hline
      \multirow{2}{*}{Method} & \multicolumn{2}{c|}{DIV2K} & \multicolumn{2}{c|}{DIV8K} & {All}\\ 
      \cline{2-6} & PSNR & LPIPS & PSNR  & LPIPS & FID \\
      \cline{1-6} GPEN & 29.129 & 0.0933 & 31.191 & 0.0703 & 3.0924\\
      \cline{1-6} FeMaSR & 29.790 & \textcolor{blue}{0.0581}  & 32.233 & 0.0416 & 1.6741 \\
      \cline{1-6} MAT & \textcolor{red}{30.124} &0.0676  & \textcolor{blue}{32.335} & \textcolor{blue}{0.0406} & \textcolor{blue}{1.2385}\\
      \cline{1-6} Ours &\textcolor{blue}{30.1151} & \textcolor{red}{0.0516
    } & \textcolor{red}{32.701} & \textcolor{red}{0.0372} & \textcolor{red}{1.1657}\\
      \hline
    \end{tabular}
    }
\end{table}

\subsection{Ablation Study}
\begin{figure}[h!]
    \centering
    \includegraphics[width=0.95\columnwidth]{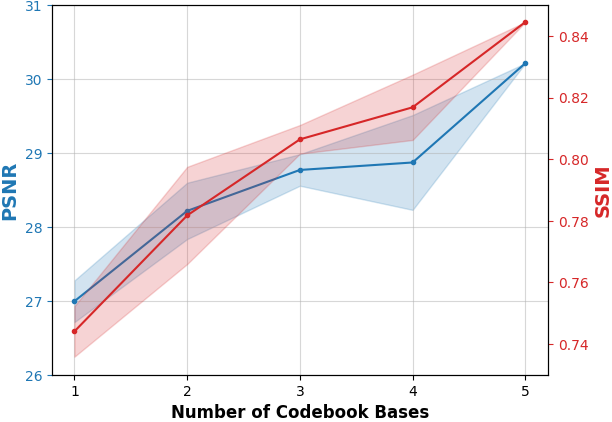}
    \caption{\textbf{Ablation Study.} PSNR and SSIM score of reconstructed results with varying number of basis codebooks. See supplementary for more result images.} 
    \label{fig:ablation}
\end{figure}
We investigate AdaCode's expressiveness given a various number of basis codebooks. We fix the five basis codebooks trained in Stage \uppercase\expandafter{\romannumeral1} and train Stage \uppercase\expandafter{\romannumeral2} with various combinations of basis codebooks. We adopt PSNR and SSIM to evaluate the expressiveness on the reconstruction task. Fig. \ref{fig:ablation} empirically shows that the adaptive codebook benefits from the bases in a large extent. Meanwhile, it also indicates that our basis codebooks are ``non-multicollinear" even if the semantic sub-datasets have overlapping patches.

\section{Conclusion and Limitation}
In this work, we propose AdaCode, a novel approach for class-agnostic image reconstruction and restoration. In particular, we train a set of class-specific basis codebooks and learn a weight map to construct an image-adaptive codebook for better image representation. Unlike previous methods that use a general codebook to represent images, our image-adaptive codebook is more flexible and suited for natural images. Extensive comparisons on image reconstruction, super-resolution, and image inpainting tasks validate our method's superiority.

Our work is a first step towards class-agnostic generative prior for arbitrary images. It has several limitations we plan to explore in future work. First, it is yet unclear how many basis codebooks and how many code entries in each codebook we need. Stage I is trained separately from Stage II\&III, which may be suboptimal. 
Second, we do not yet incorporate high-level explicit semantic information such as semantic segmentation into the framework, which may be also useful for general image restoration tasks. 
Finally, it would be interesting to extend AdaCode for high-dimensional visual appearance, such as videos and multi-spectral images.

{\small
\bibliographystyle{ieee_fullname}
\bibliography{egbib}

\begin{thebibliography}{10}\itemsep=-1pt

\bibitem{Agustsson_2017_CVPR_Workshops}
Eirikur Agustsson and Radu Timofte.
\newblock NTIRE 2017 Challenge on Single Image Super-Resolution: Dataset and
  Study.
\newblock In {\em Proceedings of the IEEE Conference on Computer Vision and
  Pattern Recognition (CVPR) Workshops}, 2017.

\bibitem{chan2021glean}
Kelvin~C.K. Chan, Xintao Wang, Xiangyu Xu, Jinwei Gu, and Chen~Change Loy.
\newblock GLEAN: Generative Latent Bank for Large-Factor Image
  Super-Resolution.
\newblock In {\em Proceedings of the IEEE/CVF Conference on Computer Vision and
  Pattern Recognition (CVPR)}, 2021.

\bibitem{chen2022real}
Chaofeng Chen, Xinyu Shi, Yipeng Qin, Xiaoming Li, Xiaoguang Han, Tao Yang, and
  Shihui Guo.
\newblock Real-World Blind Super-Resolution via Feature Matching with Implicit
  High-Resolution Priors.
\newblock In {\em Proceedings of ACM International Conference on Multimedia
  (MM)}, 2022.

\bibitem{chen2020simple}
Ting Chen, Simon Kornblith, Mohammad Norouzi, and Geoffrey Hinton.
\newblock A Simple Framework for Contrastive Learning of Visual
  Representations.
\newblock In {\em Proceedings of International Conference on Machine Learning
  (ICML)}, 2020.

\bibitem{deng2018uv}
Jiankang Deng, Shiyang Cheng, Niannan Xue, Yuxiang Zhou, and Stefanos
  Zafeiriou.
\newblock UV-GAN: Adversarial Facial UV Map Completion for Pose-Invariant Face
  Recognition.
\newblock In {\em Proceedings of the IEEE Conference on Computer Vision and
  Pattern Recognition (CVPR)}, 2018.

\bibitem{dong2022incremental}
Qiaole Dong, Chenjie Cao, and Yanwei Fu.
\newblock Incremental Transformer Structure Enhanced Image Inpainting With
  Masking Positional Encoding.
\newblock In {\em Proceedings of the IEEE/CVF Conference on Computer Vision and
  Pattern Recognition (CVPR)}, 2022.

\bibitem{esser2021taming}
Patrick Esser, Robin Rombach, and Bjorn Ommer.
\newblock Taming Transformers for High-Resolution Image Synthesis.
\newblock In {\em Proceedings of the IEEE/CVF Conference on Computer Vision and
  Pattern Recognition (CVPR)}, 2021.

\bibitem{fadili2009inpainting}
Mohamed-Jalal Fadili, J-L Starck, and Fionn Murtagh.
\newblock Inpainting and Zooming Using Sparse Representations.
\newblock {\em The Computer Journal}, 52(1):64--79, 2009.

\bibitem{fu2022kxnet}
Jiahong Fu, Hong Wang, Qi Xie, Qian Zhao, Deyu Meng, and Zongben Xu.
\newblock KXNet: A Model-Driven Deep Neural Network for Blind
  Super-Resolution.
\newblock In {\em Proceedings of European Conferences on Computer Vision
  (ECCV)}, 2022.

\bibitem{goodfellow2014generative}
Ian Goodfellow, Jean Pouget-Abadie, Mehdi Mirza, Bing Xu, David Warde-Farley,
  Sherjil Ozair, Aaron Courville, and Yoshua Bengio.
\newblock Generative Adversarial Nets.
\newblock In {\em Proceedings of Advances in Neural Information Processing
  Systems (NeurIPS)}, 2014.

\bibitem{gu2022vector}
Shuyang Gu, Dong Chen, Jianmin Bao, Fang Wen, Bo Zhang, Dongdong Chen, Lu Yuan,
  and Baining Guo.
\newblock Vector Quantized Diffusion Model for Text-to-Image Synthesis.
\newblock In {\em Proceedings of the IEEE/CVF Conference on Computer Vision and
  Pattern Recognition (CVPR)}, 2022.

\bibitem{gu2019div8k}
Shuhang Gu, Andreas Lugmayr, Martin Danelljan, Manuel Fritsche, Julien Lamour,
  and Radu Timofte.
\newblock DIV8K: DIVerse 8K Resolution Image Dataset.
\newblock In {\em Proceedings of the IEEE/CVF International Conference on
  Computer Vision (ICCV) Workshops}, 2019.

\bibitem{guo2021image}
Xiefan Guo, Hongyu Yang, and Di Huang.
\newblock Image Inpainting via Conditional Texture and Structure Dual
  Generation.
\newblock In {\em Proceedings of the IEEE/CVF International Conference on
  Computer Vision (ICCV)}, 2021.

\bibitem{heusel2017gans}
Martin Heusel, Hubert Ramsauer, Thomas Unterthiner, Bernhard Nessler, and Sepp
  Hochreiter.
\newblock GANs Trained by a Two Time-Scale Update Rule Converge to a Local Nash
  Equilibrium.
\newblock In {\em Proceedings of Advances in Neural Information Processing
  Systems (NeurIPS)}, 2017.

\bibitem{hong2018learning}
Seunghoon Hong, Xinchen Yan, Thomas~S Huang, and Honglak Lee.
\newblock Learning Hierarchical Semantic Image Manipulation through Structured
  Representations.
\newblock In {\em Proceedings of Advances in Neural Information Processing
  Systems (NeurIPS)}, 2018.

\bibitem{ji2020real}
Xiaozhong Ji, Yun Cao, Ying Tai, Chengjie Wang, Jilin Li, and Feiyue Huang.
\newblock Real-World Super-Resolution via Kernel Estimation and Noise
  Injection.
\newblock In {\em Proceedings of the IEEE/CVF Conference on Computer Vision and
  Pattern Recognition (CVPR) Workshops}, 2020.

\bibitem{johnson2016perceptual}
Justin Johnson, Alexandre Alahi, and Li Fei-Fei.
\newblock Perceptual Losses for Real-Rime Style Transfer and Super-Resolution.
\newblock In {\em Proceedings of European Conferences on Computer Vision
  (ECCV)}, 2016.

\bibitem{karras2017progressive}
Tero Karras, Timo Aila, Samuli Laine, and Jaakko Lehtinen.
\newblock Progressive Growing of GANs for Improved Quality, Stability, and
  Variation.
\newblock In {\em Proceedings of International Conference on Learning
  Representations (ICLR)}, 2018.

\bibitem{li2022mat}
Wenbo Li, Zhe Lin, Kun Zhou, Lu Qi, Yi Wang, and Jiaya Jia.
\newblock MAT: Mask-Aware Transformer for Large Hole Image Inpainting.
\newblock In {\em Proceedings of the IEEE/CVF Conference on Computer Vision and
  Pattern Recognition (CVPR)}, 2022.

\bibitem{liang2021swinir}
Jingyun Liang, Jiezhang Cao, Guolei Sun, Kai Zhang, Luc Van~Gool, and Radu
  Timofte.
\newblock SwinIR: Image Restoration Using Swin Transformer.
\newblock In {\em Proceedings of the IEEE/CVF International Conference on
  Computer Vision (ICCV) Workshops}, 2021.

\bibitem{liang2021flow}
Jingyun Liang, Kai Zhang, Shuhang Gu, Luc Van~Gool, and Radu Timofte.
\newblock Flow-Based Kernel Prior With Application to Blind Super-Resolution.
\newblock In {\em Proceedings of the IEEE/CVF Conference on Computer Vision and
  Pattern Recognition (CVPR)}, 2021.

\bibitem{lim2017enhanced}
Bee Lim, Sanghyun Son, Heewon Kim, Seungjun Nah, and Kyoung Mu~Lee.
\newblock Enhanced Deep Residual Networks for Single Image Super-Resolution.
\newblock In {\em Proceedings of the IEEE/CVF Conference on Computer Vision and
  Pattern Recognition (CVPR) Workshops}, 2017.

\bibitem{liu20224d}
Chengxu Liu, Huan Yang, Jianlong Fu, and Xueming Qian.
\newblock 4D LUT: Learnable Context-Aware 4D Lookup Table for Image
  Enhancement.
\newblock {\em arXiv preprint arXiv:2209.01749}, 2022.

\bibitem{liu2018image}
Guilin Liu, Fitsum~A Reda, Kevin~J Shih, Ting-Chun Wang, Andrew Tao, and Bryan
  Catanzaro.
\newblock Image Inpainting for Irregular Holes Using Partial Convolutions.
\newblock In {\em Proceedings of the European conference on computer vision
  (ECCV)}, 2018.

\bibitem{liu2020rethinking}
Hongyu Liu, Bin Jiang, Yibing Song, Wei Huang, and Chao Yang.
\newblock Rethinking Image Inpainting via a Mutual Encoder-Decoder with Feature
  Equalizations.
\newblock In {\em Proceedings of European Conferences on Computer Vision
  (ECCV)}, 2020.

\bibitem{liu2022reduce}
Qiankun Liu, Zhentao Tan, Dongdong Chen, Qi Chu, Xiyang Dai, Yinpeng Chen,
  Mengchen Liu, Lu Yuan, and Nenghai Yu.
\newblock Reduce Information Loss in Transformers for Pluralistic Image
  Inpainting.
\newblock In {\em Proceedings of the IEEE/CVF Conference on Computer Vision and
  Pattern Recognition (CVPR)}, 2022.

\bibitem{menon2020pulse}
Sachit Menon, Alexandru Damian, Shijia Hu, Nikhil Ravi, and Cynthia Rudin.
\newblock PULSE: Self-Supervised Photo Upsampling via Latent Space Exploration
  of Generative Models.
\newblock In {\em Proceedings of the IEEE/CVF Conference on Computer Vision and
  Pattern Recognition (CVPR)}, 2020.

\bibitem{nazeri2019edgeconnect}
Kamyar Nazeri, Eric Ng, Tony Joseph, Faisal~Z Qureshi, and Mehran Ebrahimi.
\newblock EdgeConnect: Generative Image Inpainting with Adversarial Edge
  Learning.
\newblock {\em arXiv preprint arXiv:1901.00212}, 2019.

\bibitem{oord2018representation}
Aaron van~den Oord, Yazhe Li, and Oriol Vinyals.
\newblock Representation Learning with Contrastive Predictive Coding.
\newblock {\em arXiv preprint arXiv:1807.03748}, 2018.

\bibitem{pan2021exploiting}
Xingang Pan, Xiaohang Zhan, Bo Dai, Dahua Lin, Chen~Change Loy, and Ping Luo.
\newblock Exploiting Deep Generative Prior for Versatile Image Restoration and
  Manipulation.
\newblock {\em IEEE Transactions on Pattern Analysis and Machine Intelligence
  (TPAMI)}, 44(11):7474--7489, 2021.

\bibitem{pathak2016context}
Deepak Pathak, Philipp Krahenbuhl, Jeff Donahue, Trevor Darrell, and Alexei~A
  Efros.
\newblock Context Encoders: Feature Learning by inpainting.
\newblock In {\em Proceedings of the IEEE/CVF Conference on Computer Vision and
  Pattern Recognition (CVPR)}, 2016.

\bibitem{rad2021benefiting}
Mohammad~Saeed Rad, Thomas Yu, Claudiu Musat, Hazim~Kemal Ekenel, Behzad
  Bozorgtabar, and Jean-Philippe Thiran.
\newblock Benefiting from Bicubically Down-sampled Images for Learning
  Real-World Image Super-Resolution.
\newblock In {\em Proceedings of the IEEE/CVF Winter Conference on Applications
  of Computer Vision (WACV)}, 2021.

\bibitem{ren2020neural}
Dongwei Ren, Kai Zhang, Qilong Wang, Qinghua Hu, and Wangmeng Zuo.
\newblock Neural Blind Deconvolution Using Deep Priors.
\newblock In {\em Proceedings of the IEEE Conference on Computer Vision and
  Pattern Recognition (CVPR)}, 2020.

\bibitem{ren2019structureflow}
Yurui Ren, Xiaoming Yu, Ruonan Zhang, Thomas~H Li, Shan Liu, and Ge Li.
\newblock StructureFlow: Image Inpainting via Structure-Aware Appearance Flow.
\newblock In {\em Proceedings of the IEEE/CVF International Conference on
  Computer Vision (ICCV)}, 2019.

\bibitem{satoshi2017globally}
Iizuka Satoshi, Simo-Serra Edgar, and Ishikawa Hiroshi.
\newblock Globally and Locally Consistent Image Completion.
\newblock {\em ACM Transactions on Graphics (TOG)}, 36(4):3073659, 2017.

\bibitem{shen2009image}
Bin Shen, Wei Hu, Yimin Zhang, and Yu-Jin Zhang.
\newblock Image Inpainting via Sparse Representation.
\newblock In {\em Proceedings of International Conference on Acoustics, Speech
  and Signal Processing (ICASSP)}, 2009.

\bibitem{simonyan2014very}
Karen Simonyan and Andrew Zisserman.
\newblock Very Deep Convolutional Networks For Large-Scale Image Recognition.
\newblock {\em arXiv preprint arXiv:1409.1556}, 2014.

\bibitem{song2018spg}
Yuhang Song, Chao Yang, Yeji Shen, Peng Wang, Qin Huang, and C-C~Jay Kuo.
\newblock SPG-Net: Segmentation Prediction and Guidance Network for Image
  Inpainting.
\newblock In {\em Proceedings of The British Machine Vision Conference (BMVC)},
  2018.

\bibitem{van2017neural}
Aaron Van Den~Oord, Oriol Vinyals, et~al.
\newblock Neural Discrete Representation Learning.
\newblock In {\em Proceedings of Advances in Neural Information Processing
  Systems (NeurIPS)}, 2017.

\bibitem{wan2021high}
Ziyu Wan, Jingbo Zhang, Dongdong Chen, and Jing Liao.
\newblock High-Fidelity Pluralistic Image Completion with Transformers.
\newblock In {\em Proceedings of the IEEE/CVF International Conference on
  Computer Vision (ICCV)}, 2021.

\bibitem{wang2021unsupervised}
Longguang Wang, Yingqian Wang, Xiaoyu Dong, Qingyu Xu, Jungang Yang, Wei An,
  and Yulan Guo.
\newblock Unsupervised Degradation Representation Learning for Blind
  Super-Resolution.
\newblock In {\em Proceedings of the IEEE/CVF Conference on Computer Vision and
  Pattern Recognition (CVPR)}, 2021.

\bibitem{wang2021real}
Xintao Wang, Liangbin Xie, Chao Dong, and Ying Shan.
\newblock Real-ESRGAN: Training Real-World Blind Super-Resolution With Pure
  Synthetic Data.
\newblock In {\em Proceedings of the IEEE/CVF International Conference on
  Computer Vision (ICCV)}, 2021.

\bibitem{wang2018recovering}
Xintao Wang, Ke Yu, Chao Dong, and Chen~Change Loy.
\newblock Recovering Realistic Texture in Image Super-Resolution by Deep
  Spatial Feature Transform.
\newblock In {\em Proceedings of the IEEE/CVF Conference on Computer Vision and
  Pattern Recognition (CVPR)}, 2018.

\bibitem{wang2018image}
Yi Wang, Xin Tao, Xiaojuan Qi, Xiaoyong Shen, and Jiaya Jia.
\newblock Image Inpainting via Generative Multi-Column Convolutional Neural
  Networks.
\newblock In {\em Proceedings of Advances in Neural Information Processing
  Systems (NeurIPS)}, 2018.

\bibitem{wang2022restoreformer}
Zhouxia Wang, Jiawei Zhang, Runjian Chen, Wenping Wang, and Ping Luo.
\newblock RestoreFormer: High-Quality Blind Face Restoration From Undegraded
  Key-Value Pairss.
\newblock In {\em Proceedings of the IEEE/CVF Conference on Computer Vision and
  Pattern Recognition (CVPR)}, 2022.

\bibitem{wei2020component}
Pengxu Wei, Ziwei Xie, Hannan Lu, Zongyuan Zhan, Qixiang Ye, Wangmeng Zuo, and
  Liang Lin.
\newblock Component Divide-and-Conquer for Real-World Image Super-Resolution.
\newblock In {\em Proceedings of European Conferences on Computer Vision
  (ECCV)}, 2020.

\bibitem{xie2021segformer}
Enze Xie, Wenhai Wang, Zhiding Yu, Anima Anandkumar, Jose~M Alvarez, and Ping
  Luo.
\newblock SegFormer: Simple and Efficient Design for Semantic Segmentation with
  Transformers.
\newblock In {\em Proceedings of Advances in Neural Information Processing
  Systems (NeurIPS)}, 2021.

\bibitem{xiong2019foreground}
Wei Xiong, Jiahui Yu, Zhe Lin, Jimei Yang, Xin Lu, Connelly Barnes, and Jiebo
  Luo.
\newblock Foreground-Aware Image Inpainting.
\newblock In {\em Proceedings of the IEEE/CVF Conference on Computer Vision and
  Pattern Recognition (CVPR)}, 2019.

\bibitem{yang2022adaint}
Canqian Yang, Meiguang Jin, Xu Jia, Yi Xu, and Ying Chen.
\newblock AdaInt: Learning Adaptive Intervals for 3D Lookup Tables on Real-Time
  Tmage Enhancement.
\newblock In {\em Proceedings of the IEEE/CVF Conference on Computer Vision and
  Pattern Recognition (CVPR)}, 2022.

\bibitem{yang2022seplut}
Canqian Yang, Meiguang Jin, Yi Xu, Rui Zhang, Ying Chen, and Huaida Liu.
\newblock SepLUT: Separable Image-Adaptive Lookup Tables for Real-Time Image
  Enhancement.
\newblock In {\em Proceedings of European Conferences on Computer Vision
  (ECCV)}, 2022.

\bibitem{yang2010image}
Jianchao Yang, John Wright, Thomas~S Huang, and Yi Ma.
\newblock Image Super-Resolution via Sparse Representation.
\newblock {\em IEEE Transactions on Image Processing (TIP)}, 19(11):2861--2873,
  2010.

\bibitem{yang2021gan}
Tao Yang, Peiran Ren, Xuansong Xie, and Lei Zhang.
\newblock GAN Prior Embedded Network for Blind Face Restoration in the Wild.
\newblock In {\em Proceedings of the IEEE/CVF Conference on Computer Vision and
  Pattern Recognition (CVPR)}, 2021.

\bibitem{yu2021vector}
Jiahui Yu, Xin Li, Jing~Yu Koh, Han Zhang, Ruoming Pang, James Qin, Alexander
  Ku, Yuanzhong Xu, Jason Baldridge, and Yonghui Wu.
\newblock Vector-Quantized Image Modeling with Improved VQGAN.
\newblock {\em Proceedings of International Conference on Learning
  Representations (ICLR)}, 2022.

\bibitem{yu2019free}
Jiahui Yu, Zhe Lin, Jimei Yang, Xiaohui Shen, Xin Lu, and Thomas~S Huang.
\newblock Free-Form Image Inpainting with Gated Convolution.
\newblock In {\em Proceedings of the IEEE/CVF International Conference on
  Computer Vision (ICCV)}, 2019.

\bibitem{zeng2020learning}
Hui Zeng, Jianrui Cai, Lida Li, Zisheng Cao, and Lei Zhang.
\newblock Learning Image-Adaptive 3D Lookup Tables for High Performance Photo
  Enhancement in Real-Time.
\newblock {\em IEEE Transactions on Pattern Analysis and Machine Intelligence
  (TPAMI)}, 44(4):2058--2073, 2020.

\bibitem{zeng2019learning}
Yanhong Zeng, Jianlong Fu, Hongyang Chao, and Baining Guo.
\newblock Learning Pyramid-Context Encoder Network for High-Quality Image
  Inpainting.
\newblock In {\em Proceedings of the IEEE/CVF Conference on Computer Vision and
  Pattern Recognition (CVPR)}, 2019.

\bibitem{zhang2021designing}
Kai Zhang, Jingyun Liang, Luc Van~Gool, and Radu Timofte.
\newblock Designing a Practical Degradation Model for Deep Blind Image
  Super-Resolution.
\newblock In {\em Proceedings of the IEEE/CVF International Conference on
  Computer Vision (ICCV)}, 2021.

\bibitem{zhang2018unreasonable}
Richard Zhang, Phillip Isola, Alexei~A Efros, Eli Shechtman, and Oliver Wang.
\newblock The Unreasonable Effectiveness of Deep Features as a Perceptual
  Metric.
\newblock In {\em Proceedings of the IEEE conference on computer vision and
  pattern recognition (CVPR)}, 2018.

\bibitem{zhao2014hyperspectral}
Yong-Qiang Zhao and Jingxiang Yang.
\newblock Hyperspectral Image Denoising via Sparse Representation and Low-Rank
  Constraint.
\newblock {\em IEEE Transactions on Geoscience and Remote Sensing},
  53(1):296--308, 2014.

\bibitem{zhou2017scene}
Bolei Zhou, Hang Zhao, Xavier Puig, Sanja Fidler, Adela Barriuso, and Antonio
  Torralba.
\newblock Scene Parsing through ADE20K Dataset.
\newblock In {\em Proceedings of the IEEE conference on computer vision and
  pattern recognition (CVPR)}, 2017.

\bibitem{zhou2022towards}
Shangchen Zhou, Kelvin~CK Chan, Chongyi Li, and Chen~Change Loy.
\newblock Towards Robust Blind Face Restoration with Codebook Lookup
  TransFormer.
\newblock In {\em Proceedings of Advances in Neural Information Processing
  Systems (NeurIPS)}, 2022.

\bibitem{zhu2014fast}
Zhiliang Zhu, Fangda Guo, Hai Yu, and Chen Chen.
\newblock Fast Single Image Super-Resolution via Self-Example Learning and
  Sparse Representation.
\newblock {\em IEEE Transactions on Multimedia (MM)}, 16(8):2178--2190, 2014.

\end{thebibliography}
}

\ifarXiv
    \foreach \x in {1,...,\numbersupplementpages}
    {
        \clearpage
        \includepdf[pages={\x}]{\supplementfilename}
    }
\fi

\end{document}


\title{Learning Image-Adaptive Codebooks for Class-Agnostic Image Restoration (Supplementary material)}

\author{Paper ID 10266}

\maketitle

\section{Semantic Grouped Classes}
\begin{figure*}
    \centering
    \includegraphics[width=\textwidth]{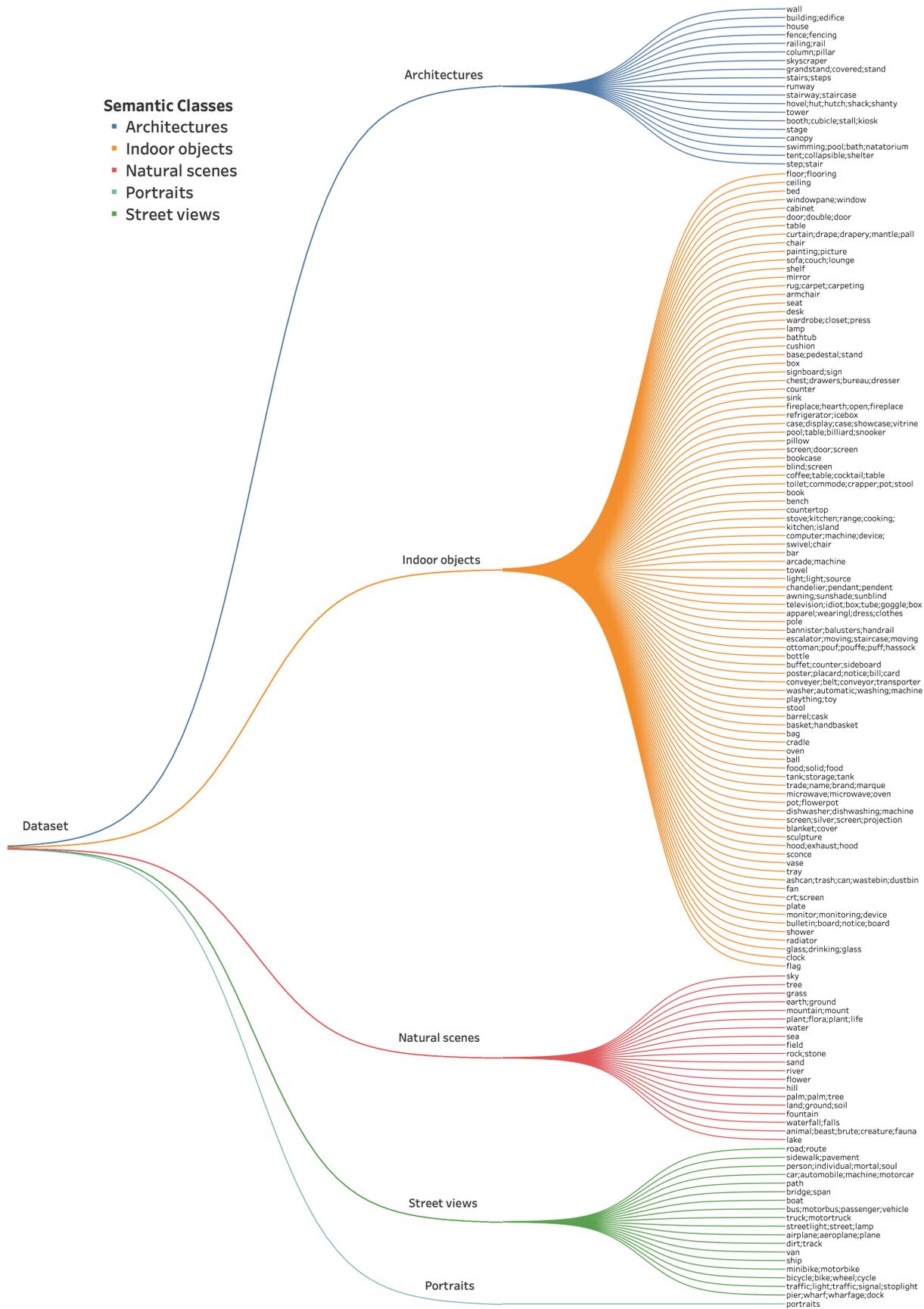}
    \caption{\textbf{Groups of 150 classes in ADE20K dataset \cite{zhou2017scene}.}} 
    \label{fig:group}
\end{figure*}

As described in the paper, we aggregate the 150 classes in ADE20K dataset \cite{zhou2017scene} into five super-classes to train our basis codebooks. The overall details of how we divide the five sub-datasets are shown in Fig. \ref{fig:group}. Although this is not a rigorous categorization, the codebook visualization in Section \textcolor{red}{3.1} empirically demonstrates that the grouping is meaningful to some extent.

\section{Network Architectures}
In conjunction with the encoder-decoder network used in our AdaCode, we elaborate on the detailed architectures of the encoder $E$, the decoder $G$, and the discriminator $D$. For $E$ and $G$, we adopt the same autoencoder as VQGAN \cite{esser2021taming} in stage \uppercase\expandafter{\romannumeral1} and the same structure as FeMaSR \cite{chen2022real} in stage \uppercase\expandafter{\romannumeral2}\&\uppercase\expandafter{\romannumeral3}.
For $D$, we adopt the same U-Net discriminator with spectral normalization as Real-ESRGAN \cite{wang2021real}.

\section{More Results}

\subsection{Ablation on Codebook Size}
We conduct experiments to empirically select the number of code entries in each basis codebook, as shown in Table. \ref{tab:Ncode}. Considering the tradeoff between performance and computation cost, we set the basis codebooks' size as $\{512, 256, 512, 256, 256\}\times 256$.

\begin{table}[h]
\caption{Stage \uppercase\expandafter{\romannumeral1} reconstruction performance on each super-class with different number of code entries. The chosen size is marked in \textcolor{red}{red}.}\label{tab:Ncode}
\centering
\footnotesize
\def\arraystretch{1.3}
\begin{tabular}{|c|c|c|c|c|}\hline
  Super-Class & Codebook Size & PSNR & SSIM & LPIPS\\ 
  \cline{1-5} \multirow{2}{*}{Architectures} & $256 \times 256$ & 24.096 & 0.667 & 0.149\\
  \cline{2-5} & \textcolor{red}{$512 \times 256$} & 24.260 & 0.684 & 0.144\\
  \cline{1-5} \multirow{2}{*}{Indoor Objects} & \textcolor{red}{$256 \times 256$} & 26.565 & 0.788 & 0.110\\
  \cline{2-5} & $512 \times 256$ &  26.630 & 0.789 & 0.110 \\
  \cline{1-5} \multirow{2}{*}{Natural Scenes} & $256 \times 256$ & 27.014 & 0.723 & 0.124 \\
  \cline{2-5} & \textcolor{red}{$512 \times 256$} & 27.693 & 0.743 & 0.110 \\
  \cline{1-5} \multirow{2}{*}{Street Views} & \textcolor{red}{$256 \times 256$} & 26.677 & 0.748 & 0.126\\
  \cline{2-5} & $512 \times 256$ & 26.937 & 0.755 & 0.127\\
  \cline{1-5} \multirow{2}{*}{Portraits} & \textcolor{red}{$256 \times 256$} &  29.914 &  0.838 & 0.097\\
  \cline{2-5} & $512 \times 256$ & 29.662 & 0.837 & 0.098\\
  \cline{1-5}
\end{tabular}

\end{table}

\subsection{Codebook Visualization}
We visualize all the codes in our five basis codebooks in Fig. \ref{fig:codebook_all}. As we discussed in Section \textcolor{red}{5}, it is yet unclear how many basis codebooks and how many code entries in each codebook we need. It is also reflected by the visualization that there might be some redundancies in the codebooks.

\subsection{Qualitative Results}
\begin{figure*}
    \centering
    \includegraphics[width=\textwidth]{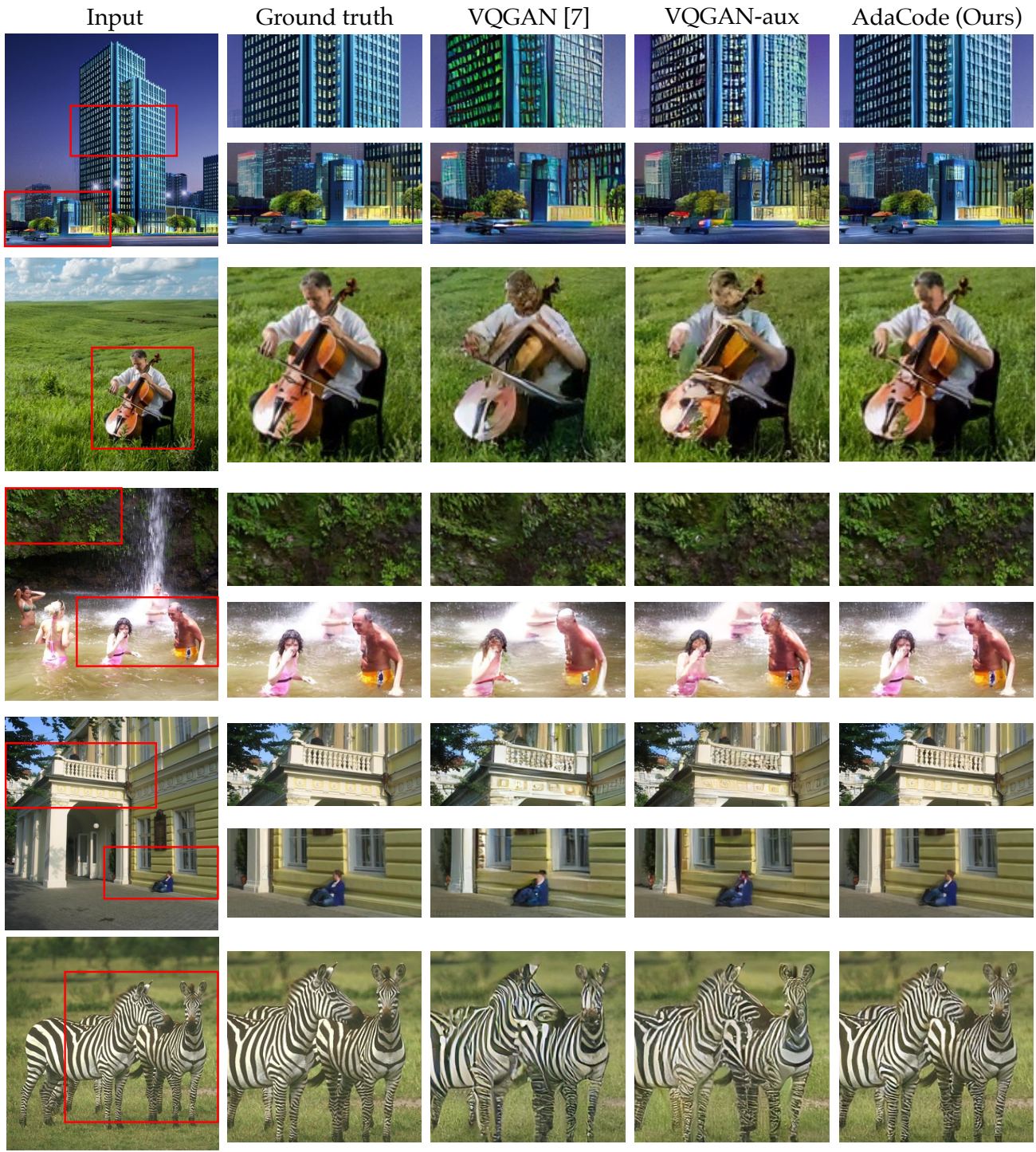}
    \caption{\textbf{More results on Image Reconstruction.}} 
    \label{fig:recon_supp}
\end{figure*}

\begin{figure*}
    \centering
    \includegraphics[width=\textwidth]{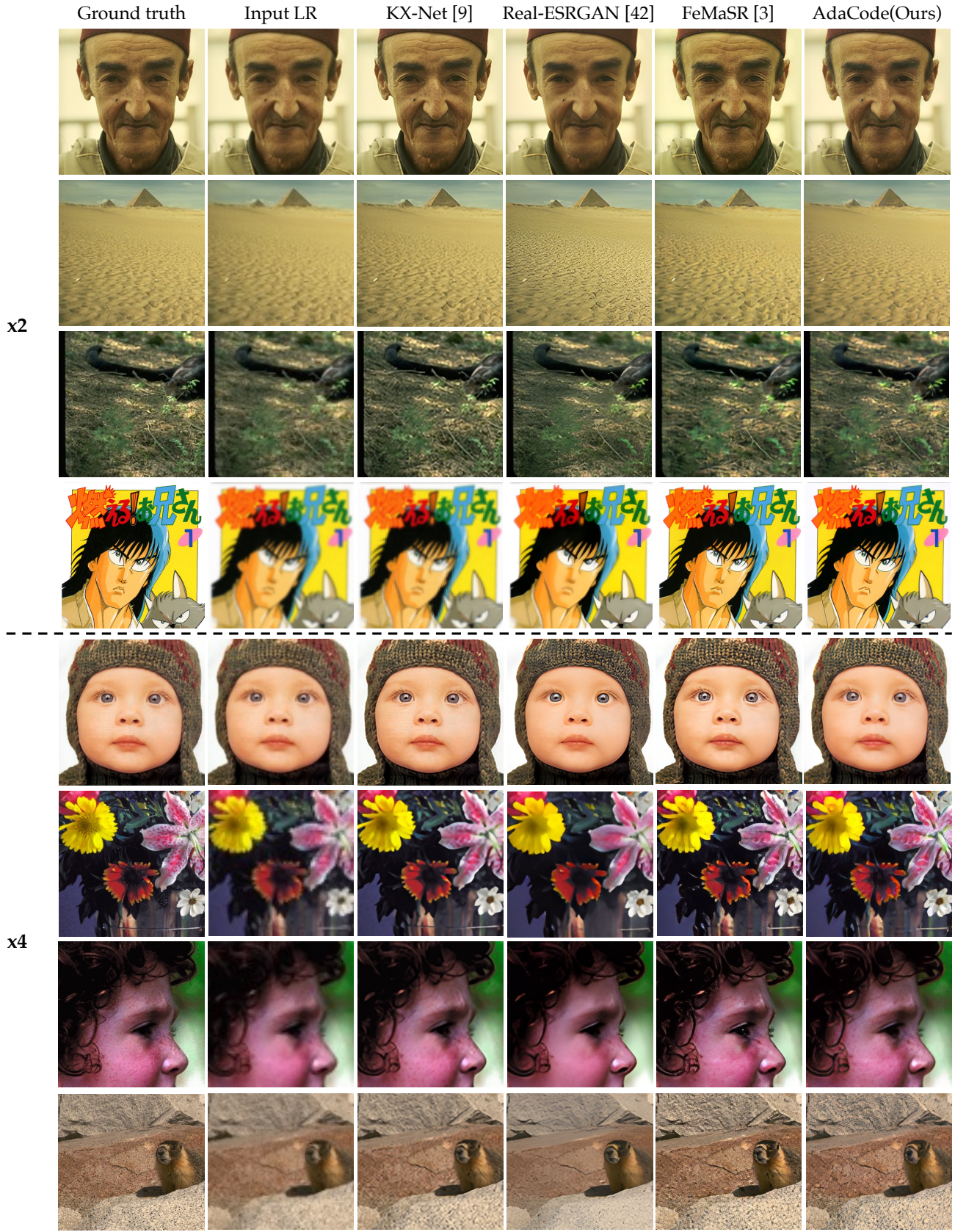}
    \caption{\textbf{More results on Super-Resolution.}} 
    \label{fig:sr_supp}
\end{figure*}

\begin{figure*}
    \centering
    \includegraphics[width=\textwidth]{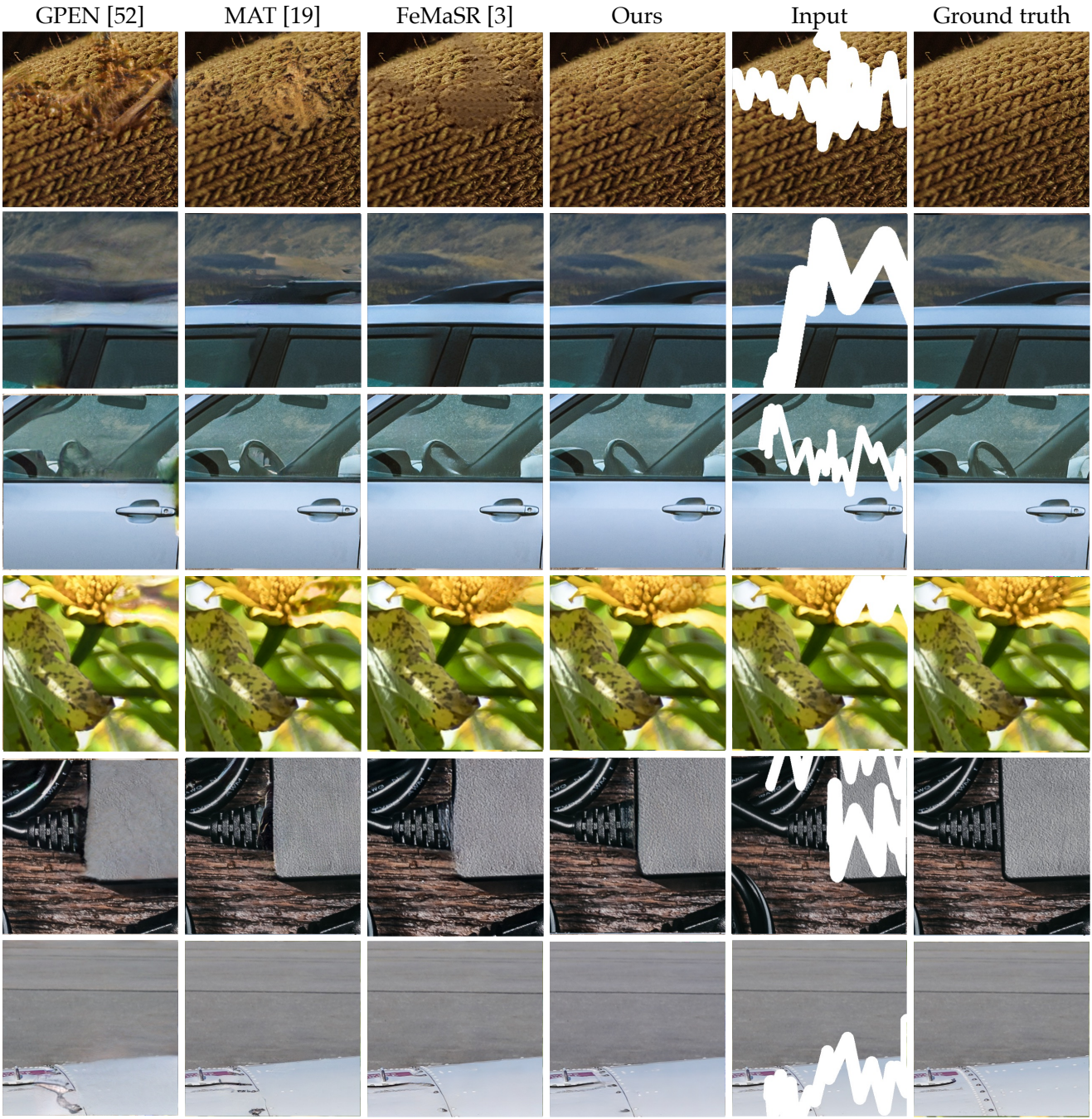}
    \caption{\textbf{More results on Image Inpainting.}} 
    \label{fig:inpainting_supp}
\end{figure*}

We show more results and comparisons for image reconstruction, super-resolution, and image inpainting in Fig. \ref{fig:recon_supp}, Fig. \ref{fig:sr_supp}, and Fig. \ref{fig:inpainting_supp}, which empirically demonstrate the effectiveness of AdaCode.

\begin{figure*}
    \centering
    \includegraphics[width=0.8\textwidth]{iccv2023AuthorKit/figures/supp/codebook.pdf}
    \caption{\textbf{Visualization of all the basis codebooks.}} 
    \label{fig:codebook_all}
\end{figure*}

{\small
\bibliographystyle{ieee_fullname}
\bibliography{egbib}
}